\relax
\documentclass[letterpaper]{article} 
\usepackage{aaai22}  
\usepackage{times}  
\usepackage{helvet} 
\usepackage{courier}  
\usepackage[hyphens]{url}  
\usepackage{graphicx} 
\usepackage{subfigure}
\usepackage{natbib}  
\usepackage{caption} 
\usepackage{color}
\usepackage{wrapfig}
\usepackage{amsmath}
\usepackage{amssymb}
\usepackage{booktabs}
\usepackage{diagbox}
\usepackage{multirow}
\usepackage{arydshln}
\usepackage[switch]{lineno}
\usepackage{tabularx}
\usepackage{stackengine}
\usepackage{xcolor}
\usepackage{adjustbox}
\usepackage{ragged2e}
\usepackage{rotating,soul}
\usepackage{tipa}
\usepackage{float}
\usepackage{bm}
\usepackage{bbding}
\usepackage{verbatim}
\usepackage{makecell}
\usepackage{overpic}
\usepackage{setspace}
\usepackage{enumitem}
\usepackage{listings}
\usepackage[noend]{algpseudocode}
\usepackage{algorithmicx,algorithm}

\usepackage{algorithmicx,algorithm}
\usepackage[noend]{algpseudocode}

\definecolor{codegreen}{rgb}{0,0.6,0}
\definecolor{codegray}{rgb}{0.5,0.5,0.5}
\definecolor{codepurple}{rgb}{0.58,0,0.82}
\definecolor{backcolour}{rgb}{0.95,0.95,0.92}

\definecolor{linkcolor}{RGB}{255,0,0}
\definecolor{urlcolor}{RGB}{255,105,180}
\definecolor{citecolor}{RGB}{0, 80, 200}
\usepackage{hyperref}
\hypersetup{colorlinks=true,linkcolor=linkcolor,urlcolor=urlcolor,citecolor=citecolor}

\definecolor{mygray}{gray}{.93}
\definecolor{mygray1}{gray}{.99}
\newcommand{\myPara}[1]{\indent\textbf{#1} \quad}
\newcommand{\myParaRelated}[1]{\noindent\textbf{#1} \quad}
\newcommand{\myParabullet}[1]{\indent\noindent$\bullet$ \textbf{#1} \quad}

\usepackage{algorithmicx,algorithm}

\def\eg{\textit{e.g.}}
\def\etc{\textit{etc}}

\lstdefinestyle{mystyle}{
    backgroundcolor=\color{backcolour},   
    commentstyle=\color{codegreen},
    keywordstyle=\color{magenta},
    numberstyle=\tiny\color{codegray},
    stringstyle=\color{codepurple},
    basicstyle=\ttfamily\footnotesize,
    breakatwhitespace=false,         
    breaklines=true,                 
    captionpos=b,                    
    keepspaces=true,                 
    numbers=left,                    
    numbersep=5pt,                  
    showspaces=false,                
    showstringspaces=false,
    showtabs=false,                  
    tabsize=2
}

\title{On Exploring and Improving Robustness of Scene Text Detection Models}
\author{
Shilian Wu\textsuperscript{\rm 1},
Wei Zhai\textsuperscript{\rm 2},
Yongrui Li\textsuperscript{\rm 2},
Kewei Wang\textsuperscript{\rm 3},
Zengfu Wang\textsuperscript{\rm 1}\thanks{Corresponding author}\\
\textsuperscript{\rm 1, \rm 2, }University of Science and Technology of China

\textsuperscript{\rm 3}Faculty of Engineering, the University of Sydney

\{wushilia,wzhai056\}@mail.ustc.edu.cn, yong.rui0@hotmail.com, kwan5498@uni.sydney.edu.au,  zfwang@ustc.edu.cn}
\begin{document}

\def\mathbi#1{\textbf{\em #1}}

\maketitle

\begin{abstract}

It is crucial to understand the robustness of text detection models with regard to extensive corruptions, since scene text detection techniques have many practical applications. For systematically exploring this problem, we propose two datasets from which to evaluate scene text detection models: ICDAR2015-C (IC15-C) and CTW1500-C (CTW-C). Our study extends the investigation of the performance and robustness of the proposed region proposal, regression and segmentation-based scene text detection frameworks. Furthermore, we perform a robustness analysis of six key components: pre-training data, backbone, feature fusion module, multi-scale predictions, representation of text instances and loss function. Finally, we present a simple yet effective data-based method to destroy the smoothness of text regions by merging background and foreground, which can significantly increase the robustness of different text detection networks. We hope that this study will provide valid data points as well as experience for future research. Benchmark, code and data will be made available at \url{https://github.com/wushilian/robust-scene-text-detection-benchmark}.


\end{abstract}

\section{Introduction}
Due to its wide practical applications such as industrial automation, image retrieval, robots navigation, and automatic driving, detecting text in natural images has already become a burgeoning research field. 

Scene text detection, which is acknowledged as the premise of reading text systems, has been driven by deep learning and massive data in the past few years and has made great progress in technology. Unlike general object detection, which requires bounding box outputs, scene text detection needs to output the explicit external polygon of the text instances. Based on this motivation, researchers have made many efforts in seeking effective contour modeling of arbitrary shape text representation, which can be roughly divided into three categories: region proposal-based frameworks \citep{xiao2020sequential,lyu2018mask}, regression-based frameworks \citep{zhu2021fourier,xue2019msr,zhang2019look} and segmentation-based frameworks \citep{liao2020real,wang2019efficient}.

Recent research focuses on improving the performance of multi-oriented text detection \citep{wang2021r} and arbitrary-shaped text detection \citep{liao2020real}. While in the real world, the deployment model requires good performance and robustness to common image corruptions. And researchers have found that model performance is prone to image corruptions \citep{azulay2019why}. Robustness has been studied for classification \citep{hendrycks2019benchmarking}, object detection \citep{michaelis2019benchmarking}, semantic segmentation \citep{kamann2020benchmarking}, \etc. But text detection requires prediction of text and non-text regions and modeling of text contours of arbitrary shapes, which is worthy of special investigations.

Inspired by \cite{hendrycks2019benchmarking}, we establish the robust scene text detection benchmarks based on ICDAR2015 \citep{karatzas2015icdar} and CTW1500 \citep{yuliang2017detecting}, consisting of two challenging datasets including ICDAR2015-C (IC15-C) and CTW1500-C (CTW-C). The benchmark datasets are constructed based on 18 different image corruptions from blur, noise, weather, digital, and geometric distortion.

To study the robustness of text detection models, we selected one representative method each from the region proposal-based, regression-based and segmentation-based frameworks. Extensive experiments show that the regression-based framework considers the overall shape of the text, and therefore has the highest robustness.

The impact of training data on performance is crucial, and many models are now pre-trained on external datasets first and then fine-tuned on the target data \citep{wang2019shape,baek2019character,feng2019textdragon}, so we investigate the effect of 2 classical pre-training datasets, SynthText \citep{gupta2016synthetic} and MLT17 \citep{nayef2017icdar2017}, on performance and robustness. We find that SynthText has a more significant improvement in robustness, while MLT17 performs better in improving performance.

The pipeline of training a scene text detector is as follows: the image goes through a backbone to extract features, then it goes through a feature fusion module to fuse features at different scales, followed by a single-scale or multi-scale predictions mechanism to predict the results, and finally the loss is calculated. Therefore we analyze the impact of these factors on the robustness.
Based on the experimental results, the following conclusions can be drawn: (1) Stronger feature extractors can improve both performance and robustness. (2) Shallow features are sensitive to noise. (3) Introducing larger receptive fields when fusing multi-scale features can improve robustness. (4) Multi-scale predictions can achieve a better trade-off between performance and robustness. (5) Considering the overall shape when modeling text instances is beneficial to improve robustness. (6) The IoU-based loss functions are more robust than L1-based.

Finally, we propose a \textbf{F}oreground and \textbf{B}ackground \textbf{Mix} (\textbf{FBMix}) augmentation method to enrich the background texture of the text area. Qualitative and quantitative experiments show that the proposed method can improve the robustness while maintaining or even improving the performance. 

Our main contributions are summarised as follows:



\begin{itemize}[leftmargin=25pt]
\item [1)] 
We propose benchmarks for text detection robustness evaluation: IC15-C and CTW-C. We systematically analyze the impact of different text detection frameworks on robustness, including region proposal-based, regression-based and segmentation-based.

\item [2)] 
We investigate the effects of pre-training data, model structure, loss function, \etc, on performance and robustness. Several interesting conclusions are drawn that will improve the performance and robustness of the text detectors.
   
\item [3)] 
We propose a simple but effective data augmentation method for text detection termed FBMix, which can significantly improves the robustness of models while maintaining the performance.

\end{itemize}

\section{Related Works}

\myParaRelated{Scene Text Detection.} Scene text detection aims to localize the polygons of text instances \citep{lyu2018multi,zhang2020deep,xie2019scene}. EAST \citep{zhou2017east} directly predicts rotated boxes of text instances, whose pipeline is flexible and simple. MaskTextSpotter \citep{lyu2018mask} detects text of arbitrary shapes by segment the instance text region proposals. PSENet \citep{wang2019shape} represents the text instance by predicting the mask of the shrunk text box, and combines the progressive scale expansion algorithm to get the final segmentation results. DBNet \citep{liao2020real} proposes a differentiable binarization (DB) module to set the thresholds for binarization adaptively. ABCNet \cite{liu2020abcnet} predicts arbitrary shaped text instances by fitting parameterized Bezier curves. TextRay \citep{wang2020textray} encodes complex geometric layouts into polar systems, proposing a single-shot anchor-free framework to jointly optimize text/non-text classification and geometric encodings regression. FCENet \citep{zhu2021fourier} models text contours in Fourier space, by which the text contours could be trained end-to-end and deployed with simple post-processing algorithm.

\myParaRelated{Benchmarking corruption robustness.} 
Several attempts have been made to deal with CNN's vulnerability to common corruption so far. \cite{hendrycks2019benchmarking} establishes a benchmark for the robustness of image classification, and applies some common visual corruptions to generate two datasets called ImageNet-C and ImageNet-P. \cite{kamann2020benchmarking}  presents a robustness study for semantic segmentation, and finds out that the backbone and some architecture properties have an impact on the improvement of robustness. An easy-to-use benchmark is proposed by \cite{michaelis2019benchmarking} to evaluate the performance of the object detection methods when the image quality decreases. Considering that the human pose estimators are easily affected by some corruptions, such as blur, noise, \etc. \cite{wang2021when} builds rigorous robust benchmarks to study the weaknesses of current excellent pose estimators.

\myParaRelated{Robustness Enhancement.} Data augmentation can greatly improve model's generalizability. However, it still remains a difficulty on how to improve model's robustness to unseen image corruptions. \cite{geirhos2018generalisation} finds that models trained under a certain type of corruption data work well under the current corruption while giving a poor performance when tested under other corruption types. In a different study, \cite{geirhos2018imagenet} trains a classification model with a stylized ImageNet dataset, reporting improved model robustness on different corruption types by ignoring texture and focusing on object shape. \cite{zhang2017mixup} proposes a data augmentation method called MixUp, which combines two paired images with their corresponding labels to generate a new image with the label. AdvMix, proposed by \cite{wang2021when}, is a data augmentation method that can effectively improve the robustness of the pose estimation model. Unfortunately, it is designed for human pose estimation and cannot be applied to text detection directly.

\section{Methodology}
\subsection{Benchmarking Robustness}
In this part, we construct benchmark datasets and evaluation metrics for text detection robustness.

\begin{figure}[t]
\centering
\begin{overpic}[width=1.0\linewidth]{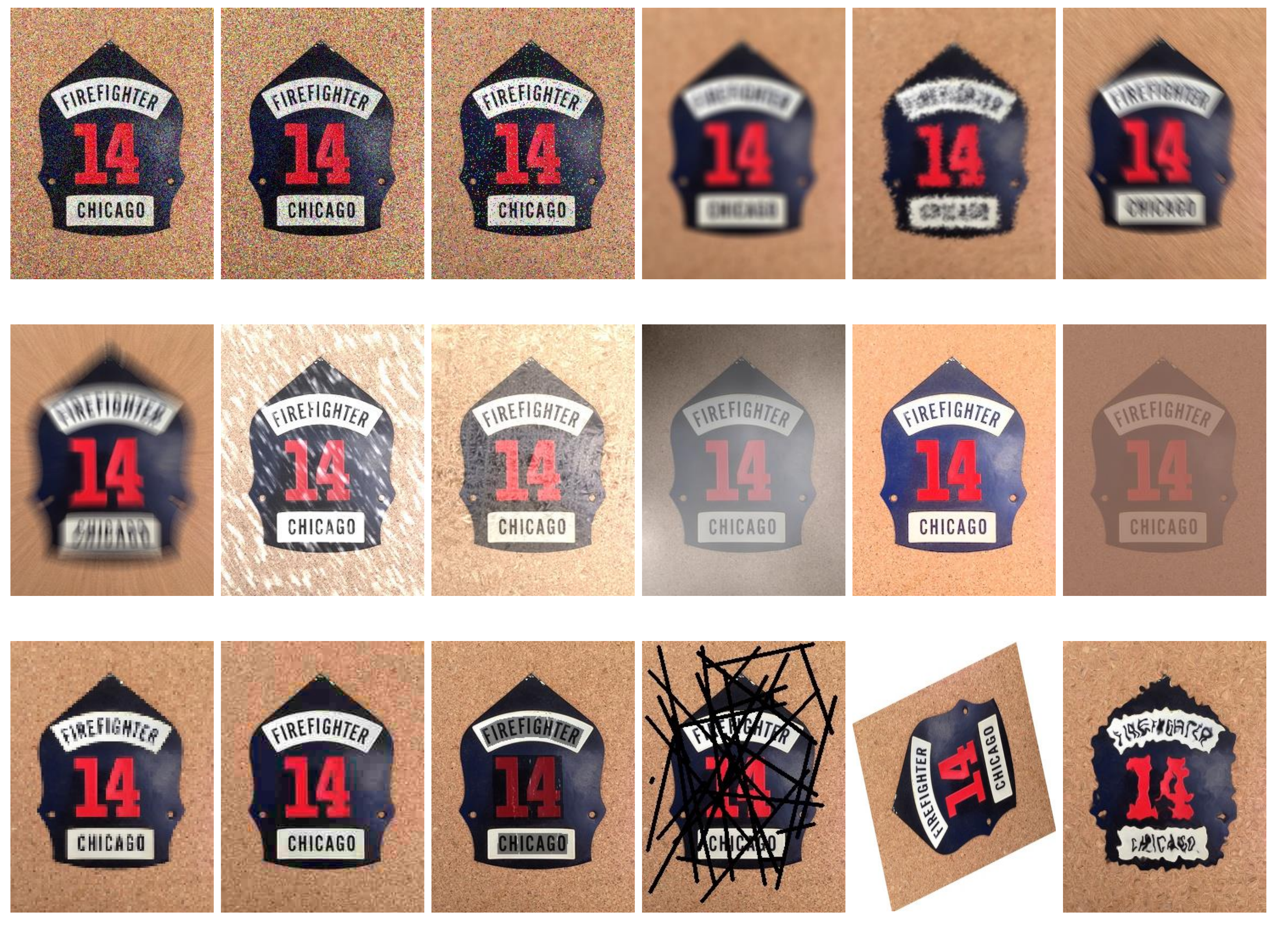}
            \put(0, 50.5){\scriptsize{Gaussian noise}}
		    \put(19.5, 50.5){\scriptsize{Shot noise}}
		    \put(33.5, 50.5){\scriptsize{Impulse noise}}
		    \put(51, 50.5){\scriptsize{Defocus blur}}
		    \put(69, 50.5){\scriptsize{Glass blur}}
		    \put(84.5, 50.5){\scriptsize{Motion blur}}
		    \put(3, 25.5){\scriptsize{Zoom blur}}
		    \put(22,25.5){\scriptsize{Snow}}
		    \put(38, 25.5){\scriptsize{Frost}}
		    \put(56,25.5){\scriptsize{Fog}}
		    \put(68.5, 25.5){\scriptsize{Brightness}}
		    \put(86, 25.5){\scriptsize{Contrast}}
		    \put(4.5, 0.5){\scriptsize{Pixelate}}
		    \put(22.5, 0.5){\scriptsize{Jpeg}}
		    \put(39, 0.5){\scriptsize{Dirty}}
		    \put(55, 0.5){\scriptsize{Lines}}
		    \put(70, 0.5){\scriptsize{Rotation}}
		    \put(88, 0.5){\scriptsize{Elastic}}
\end{overpic}
\caption{Visualization of images from our benchmark datasets, which  consists of 18 types of corruptions from noise, blur, weather, digital and geometry categories. There are five levels of severity for each type of corruption, leading to 90 distinct corruptions. }
\label{fig:corruptions}
\end{figure}
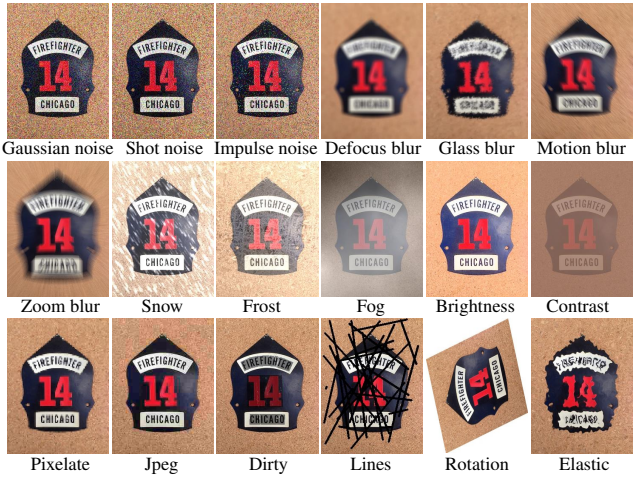

\myPara{Benchmark Datasets.}
The robust scene text detection benchmark is composed of two benchmark datasets: IC15-C, CTW-C. They are constructed by applying 18 different types of image corruption with five severity levels. Fifteen of these corruptions are from the ImageNet-C dataset, and the other three are our proposed corruptions for text images. Some example images are shown in Fig. \ref{fig:corruptions}. Considering that the real-world text samples are likely to be covered with stains or painted by others, we propose two kinds of corruption: dirty and lines. In addition, we also rotate the image at different angles to obtain a new type of rotation corruption. Ultimately, all corruptions can be divided into five types: noise, blur, weather, digital, and geometry.

\myPara{Evaluation Metrics.}
For text detection tasks, precision, recall and F-measure are the standard evaluation metrics. The F-measure is calculated from precision and recall, which reflects performance in a balanced way. So we apply F-measure as a performance metric for every model, and F is used to denote F-measure. According to \cite{michaelis2019benchmarking}, we evaluate model robustness by \textbf{m}ean \textbf{P}erformance under \textbf{C}orruption (\textbf{mPC}): $mPC=\frac{1}{N_c}\sum\limits_{c=1}^{N_c}\frac{1}{N_s}\sum\limits_{s=1}^{N_s}F_{c,s}.$ Here, $F_{c,s}$ denotes the F-measure of the model under corruption type $c$ with severity level $s$. $Nc = 18$ and $Ns = 5$
are the numbers of corruption types and severity levels, respectively. To measure the degradation of performance under different corruptions, we define the \textbf{r}elative \textbf{P}erformance under \textbf{C}orruption (\textbf{rPC}) as follows: $rPC=\frac{mPC}{F_{clean}}.$ Here, $F_{clean}$ represents the performance on a clean dataset.

\subsection{Exploring Robustness}
To analyze which factors affect the robustness of the text detection algorithm, we study the influence of frameworks, pre-training data, backbone, feature fusion module, multi-scale predictions, representation of text instances and loss function on the robustness. Then we propose a simple and effective data augmentation method, which is demonstrated by qualitative and quantitative experiments.

\begin{figure}[t]
\centering
\begin{overpic}[width=1.0\linewidth]{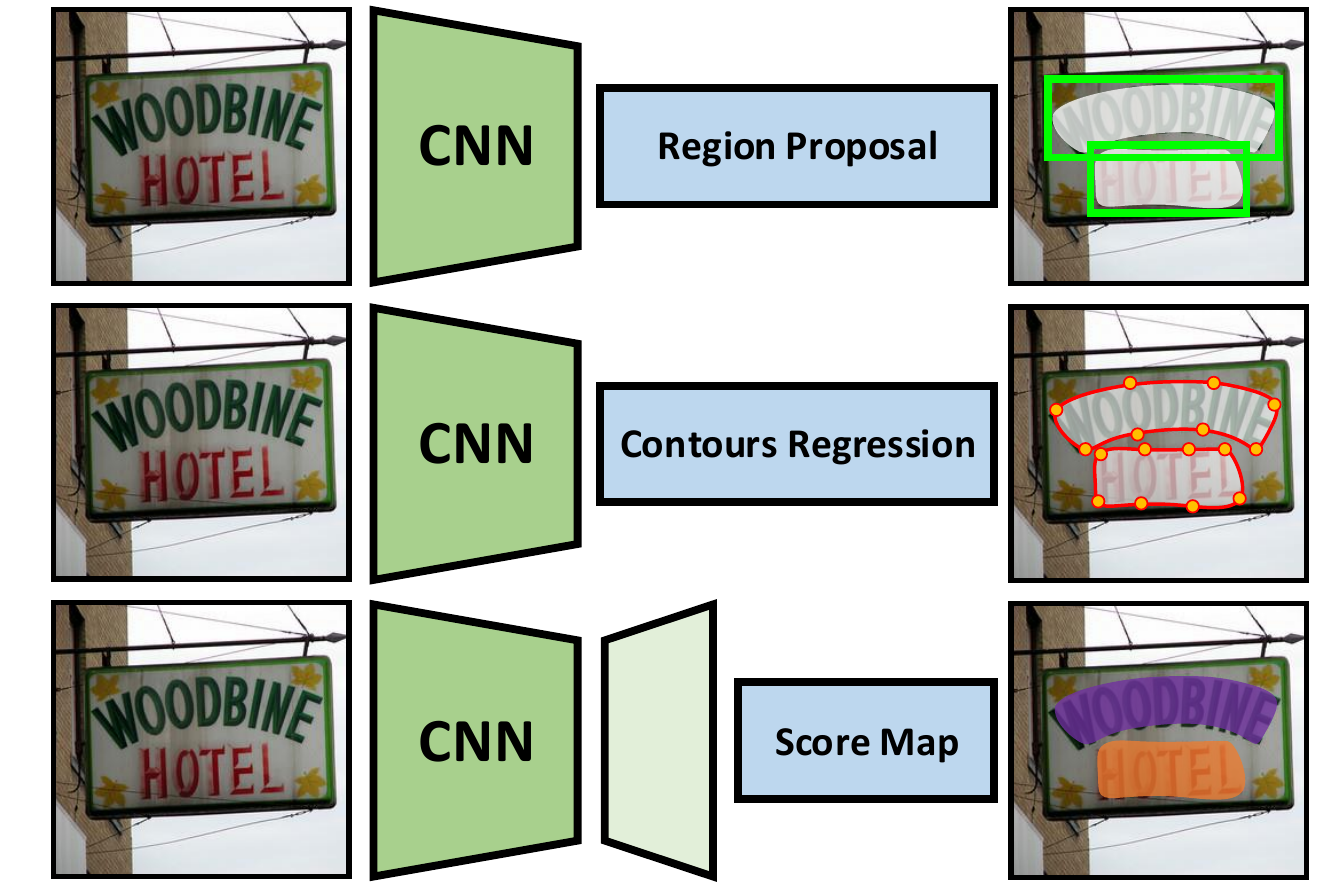}
        \put(-0.8, 57){\small{(a)}}
        \put(-0.8, 33){\small{(b)}}
        \put(-0.8, 10){\small{(c)}}
\end{overpic}
\caption{Text detection frameworks: (a) Region proposal-based. (b) Regression-based. (c) Segmentation-based.}
\label{fig:framework}
\end{figure}

\myPara{Frameworks for text detection.} Text detection models can be divided into three types as shown in Fig. \ref{fig:framework}. \textbf{Region proposal-based:} inspired by RCNN-based detection frameworks, these methods predict the contour points or masks of the text instances based on region proposals. And many region proposal-based text detection methods are derived from MaskRCNN \citep{he2017mask} (MSRCNN), so we use MSRCNN to detect text of arbitrary shapes by segment the instance text region proposals (see Fig. \ref{fig:framework} (a)). \textbf{Regression-based:} inspired by object detection, these methods rely on a contour regression-based framework. FCENet models text contours in Fourier space (see Fig. \ref{fig:framework} (b)). \textbf{Segmentation-based:} inspired by semantic segmentation, these methods use per-pixel masks to encode text instances. PSENet predicts text kernels at different scales and finally merges the text kernels by a post-processing algorithm to obtain the results (see Fig. \ref{fig:framework} (c)).

\begin{figure}[t]
\centering
\begin{overpic}[width=1.0\linewidth,]{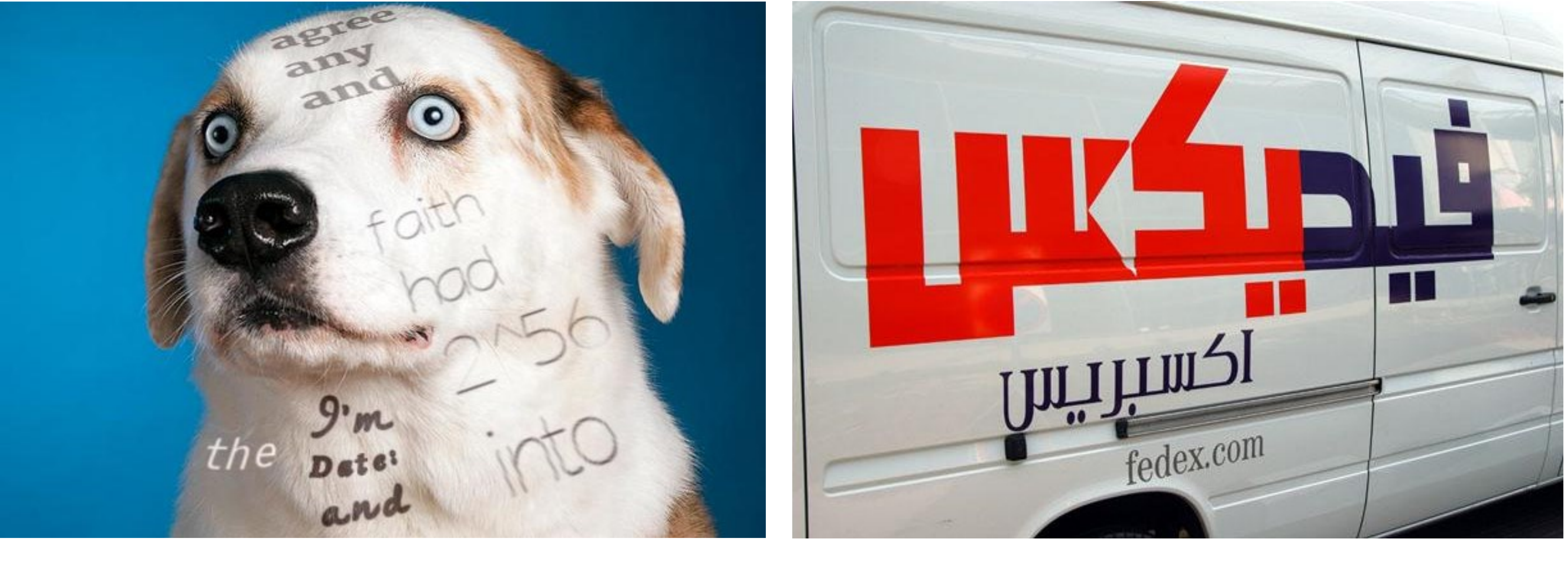}
    \put(16, -0.5){\footnotesize{SynthText}}
    \put(2, -3.3){\scriptsize{\cite{gupta2016synthetic}}}
    \put(69,-0.5){\footnotesize{MLT17}}
    \put(64,-3.3){\scriptsize{\cite{nayef2017icdar2017}}}
\end{overpic}
\caption{Examples of pre-training data.}
\label{fig:pretraindata}
\end{figure}

\myPara{Pre-training Data.}
There are many methods that first pre-train the model using synthetic data (\eg{ SynthText \citep{gupta2016synthetic}}) or real large-scale data (\eg{ MLT17 \citep{nayef2017icdar2017}} ). Then models are finetuned on the target dataset. Some examples of these two datasets are shown in the Fig. \ref{fig:pretraindata}, SynthText does not take semantic information into account, so text may appear on arbitrary backgrounds. Different pre-training data may lead to different performances, but current methods usually ignore this compared with other methods. And we also want to investigate the effect of different pre-training data on the robustness of the model, so we first pre-train three types of text detection models with different pre-training data and then finetune it on the IC15 and CTW datasets, respectively. Finally, the three types of text detection algorithms are tested separately, and the test results can objectively reflect the impact of pre-training data on performance and robustness.

\myPara{BackBone.}
While most previous text detection models use ResNet50 \citep{he2016deep} or VGG16 \citep{simonyan2015very} to extract visual features, there are now newer backbones that have been proven to improve performance in object detection tasks. We utilize VGG16, ResNet50, ResNeXt50 \citep{xie2017aggregated}, RegNetX-4.0GF \citep{radosavovic2020designing} as backbone. Each backbone is applied to three text detection models, trained on the IC15 and CTW datasets respectively.

\myPara{Feature Fusion.} 
\begin{figure*}[t]
\centering
\begin{overpic}[width=0.989\linewidth]{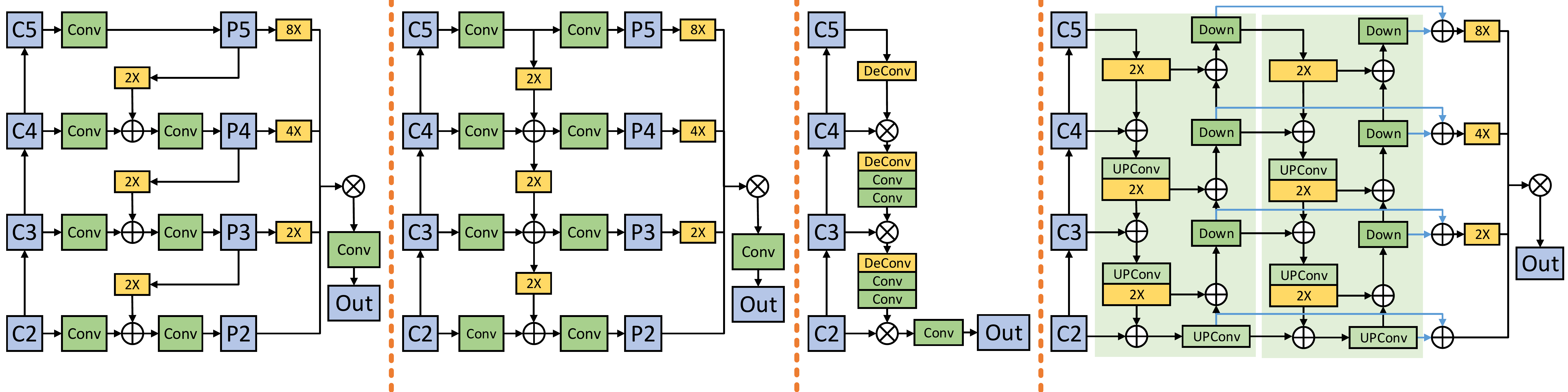}
		    \put(8, 0){\small{(a) FPNF}}
		    \put(34.5, 0){\small{(b) FPNC}}
		    \put(53.5, 0){\small{(c) FPN\_UNET}}
		    \put(79, 0){\small{(d) FFM}}
\end{overpic}
\caption{Feature fusion modules. (a) ``FPNF'' comes from PSENet \citep{wang2019shape}. (b) ``FPNC'' comes from DBNet \citep{liao2020real}. (c) ``FPN\_UNET'' comes from TextSnake \citep{long2018textsnake}. (d) ``FFM'' comes from PAN \citep{wang2019efficient}.}
\label{fig:fpn}
\end{figure*}
Scene text detectors usually fuse the multi-scale features generated by the backbone and then use the fused features for subsequent processing. For the segmentation-based framework, PSENet \citep{wang2019shape}, PAN \citep{wang2019efficient}, DBNet \cite{liao2020real} and TextSnake \citep{long2018textsnake} use different feature fusion modules respectively as shown in Fig. \ref{fig:fpn}. We want to investigate the effect of this module on the robustness of the model. Therefore, we embed the different feature fusion modules into the PSENet and perform training and testing. Two types of feature pyramid networks (FPN), PAFPN \citep{liu2018path} and CARAFE \citep{wang2019carafe}, have been shown to improve performance in object detection tasks, and region proposal-based and regression-based frameworks originate from object detection, so we apply these two types of FPN on MSRCNN and FCENet, then conduct experiments.

\myPara{Multi Scale Predictions.}
In general, the input features are passed through FPN to obtain multi-scale features. Some methods fuse different levels of features into a single scale feature and perform single-scale prediction based on this feature \citep{zhou2017east,wang2020r}. There are also some methods conduct prediction on the multi-scale features \citep{zhu2021fourier,wang2020textray,liu2020abcnet}. Such as FCENet, which makes predictions on the feature maps P3, P4, and P5 of FPN. We modify FCENet to predict only on P3, P4, or P5, respectively, and then compare the performance with the predictions of all scale features.

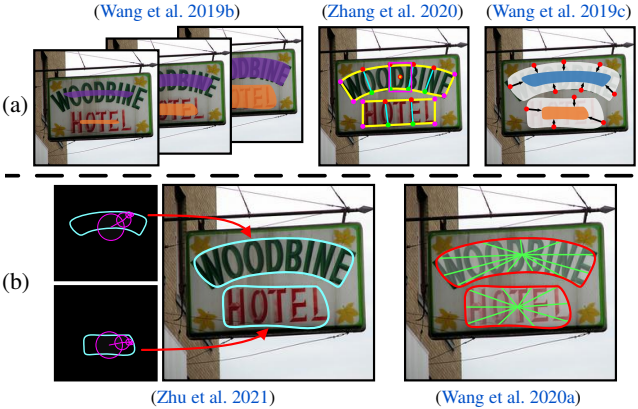
\begin{figure}[t]
\centering
\begin{overpic}[width=1.0\linewidth]{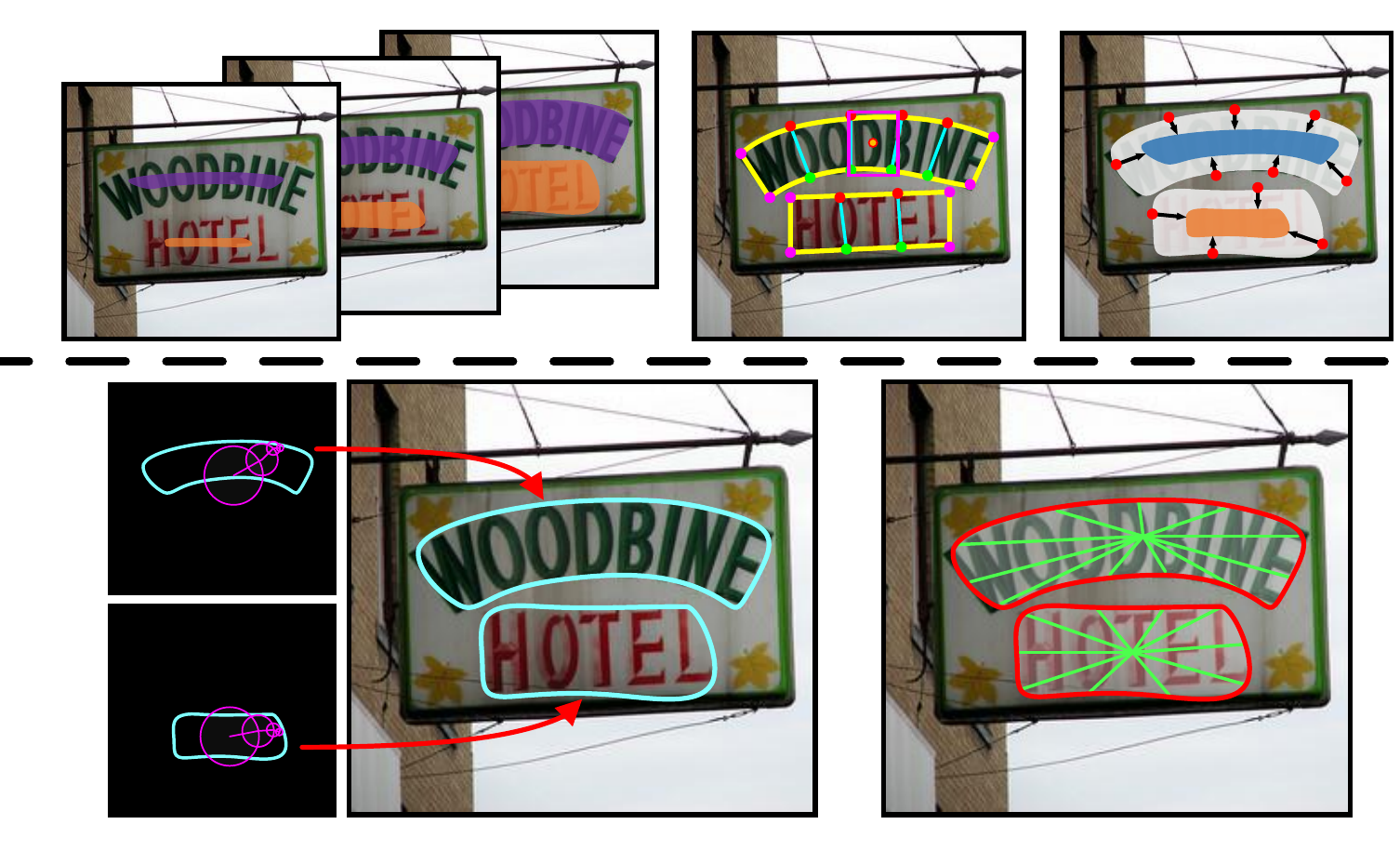}
        \put(-0.5, 45){\small{(a)}}
        \put(-0.5, 17){\small{(b)}}
        
        \put(23, -1){\scriptsize\cite{zhu2021fourier}}
        \put(68, -1){\scriptsize\cite{wang2020textray}}
        
        \put(14, 60){\scriptsize\cite{wang2019shape}}
        \put(50, 60){\scriptsize\cite{zhang2020deep}}
        \put(76, 60){\scriptsize\cite{wang2019efficient}}
\end{overpic}
\caption{Representation of text instances. (a) Three segmentation-based representations: different scale of text kernels, series of small components and text region with similar vectors. (b) Two regression-based representations: Fourier domain and Polar coordinate domain.}
\label{fig:representation}
\end{figure}
\myPara{Representation of Text Instances.} The focus of recent research has shifted from multi-oriented text detection to arbitrary-shaped text detection, and scholars are investigating how to represent arbitrary-shaped text better. PSENet \citep{wang2019shape} uses text kernels of different scales to represent text instances; DRRG \citep{zhang2020deep} uses local components to represent a region of a text instance and finally merges the components to get the whole text; PAN \citep{wang2019efficient} uses similar vectors to distinguish between different text instances; FCENet \citep{zhu2021fourier} represents text contours in Fourier space, and TextRay \citep{wang2020textray} uses multiple rays to represent text contours in polar space. These five representations are shown in the Fig. \ref{fig:representation}, and we have also studied and analyzed them systematically. For the segmentation-based framework, we fix the backbone as ResNet50, set the feature fusion module as FPN\_UNET \citep{long2018textsnake}, and train the model on the CTW dataset using three prediction heads: PSENet, DRRG, and PAN. For the regression-based framework, we fix the backbone as ResNet50 with DCN \citep{zhu2019deformable}, set the feature fusion module as FPN, and use the two prediction heads from FCENet and TextRay to train the model on the CTW dataset.

\myPara{Loss Function.}
The performance of text detection method is usually evaluated based on Intersection over Union (IoU) for text detection tasks. However, the general distance loss does not reflect the IoU between the predicted bounding box and the ground truth. Therefore, we conduct experiments on four loss functions (L1 loss, Smooth L1 loss, GIoU loss \citep{rezatofighi2019generalized}, DIoU loss \citep{zheng2020distance}) to investigate their influences on the performance and robustness of the scene text detection model.

\subsection{Foreground and Background Mix (FBMix)}

\begin{algorithm}[t]
\caption{FBMix in a Python-like style}
\begin{lstlisting}[language=Python]
import cv2
    
def FBMix(fgimg, bgimg, alpha=0.5):   
    oh, ow, c = bgimg.shape() #background
    dh, dw, c = fgimg.shape() #foreground
    if oh! = dh or ow! = dw:
        bgimg = cv2.resize(bgimg, (dw, dh))
    result = fgimg * alpha + bgimg * (1 - alpha) 
    return result
\end{lstlisting}
\label{algorithm:fgmix}
\end{algorithm}
\begin{figure}[t]
\centering
\begin{overpic}[width=1.0\linewidth,]{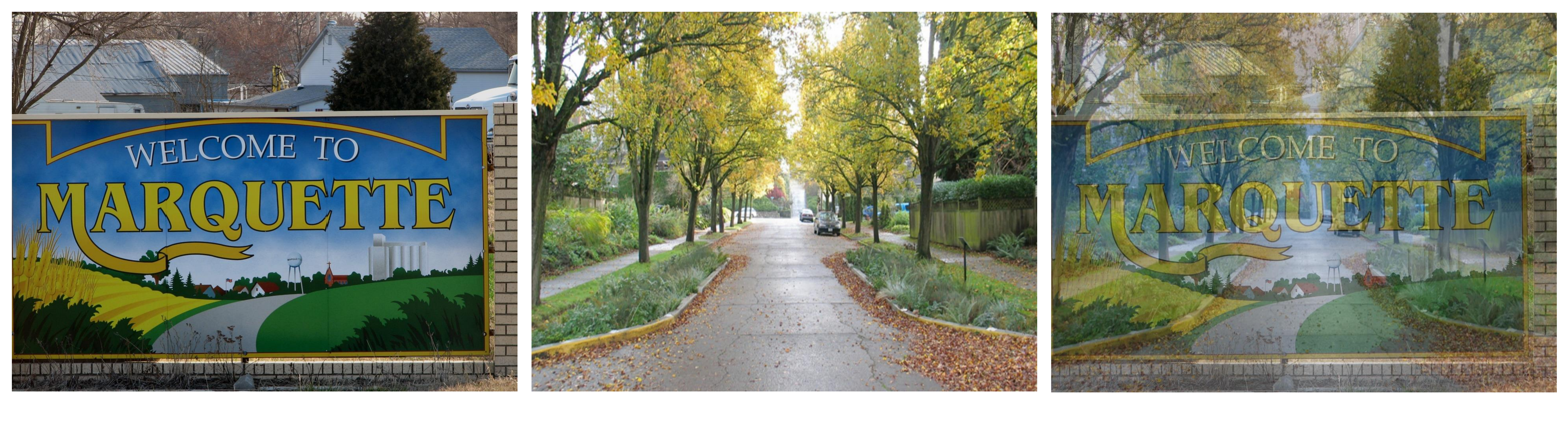}
            \put(7.9, -1.){\small{Foreground}}
            \put(41.4, -1.){\small{Background}}
            \put(77.5, -1.){\small{FBMix}}
\end{overpic}
\caption{A realization of FBMix, combine foreground image and background image together to get a new image.}
\label{fig:fgmix}
\end{figure}
In general, text generally appears in smooth background areas and has periodic consistency within the text, and this prior knowledge may fail in some cases. For the model to learn the essential features of the text while ignoring prior knowledge as much as possible, we propose the data augmentation method of FBMix. FBMix randomly selects a background image and does a weighted fusion with the training image by the coefficient $\alpha$ as shown in Fig. \ref{fig:fgmix}. A concrete account of the algorithm is given in Algorithm \ref{algorithm:fgmix}. This approach will destroy the smoothness of the text area and significantly improve the richness of the texture in the background area of the text. In this paper, $\alpha$ is set to 0.5.

\section{Experiments and Analyses}

\subsection{Implementation Details}
Based on the open-source toolbox MMOCR \citep{kuang2021mmocr},
we re-trained the models mentioned in the previous section to make a fair comparison. In addition, the three types of text detection models require different numbers of iterations to converge. For MSRCNN, PSENet and FCENet, we train 160, 600, 1500 epochs on the corresponding dataset, respectively. All experiments are conducted on a computer with eight RTX 3090 GPUs.

\subsection{Discussion of Robustness Study}
\begin{table}[t]
\centering
\small
\renewcommand{\arraystretch}{1.}
\renewcommand{\tabcolsep}{4.3pt}
\begin{tabular}{c||c|ccccc|c|c} 
\hline
\textbf{$\mathcal{M}$} & \textbf{Clean} & \textbf{N.} & \textbf{B.}  & \textbf{W.} & \textbf{D.} & \textbf{G.}  & \textbf{mPC} & \textbf{rPC}   \\ 
\hline
Regi.  & 82.5 & 16.7 & 26.2 & 61.6   & 62.4   & 68.0     & 47.2   & 57.3   \\
Regr.    & \textbf{84.9} & 19.6 & \textbf{30.1} & \textbf{65.9}   & \textbf{67.3}   & \textbf{76.8}    & \textbf{51.9}   & \textbf{61.1}    \\
 Seg.    & 80.7 & \textbf{21.6} & 26.9 & 59.5   & 63.2   & 67.8     & 48.1   & 59.6    \\
\hline
\end{tabular}
\caption{Results of different frameworks on IC15. ``Regi.'', ``Regr.'' and ``Seg.'' denotes region proposal-based framework, regression-based framework and segmentation-based framework, respectively. ``N.'', ``B.'', ``W.'', ``D.'' and ``G.'' denotes Noise, Blur, Weather, Digital and Geometry.}
\label{table:framework}
\end{table}
In this part, we conduct experiments and analyses on the factors affecting the robustness of the text detection model.

\myParabullet{Performance $w.r.t$ Frameworks.} The results of different text detection frameworks on IC15 are shown in the Table \ref{table:framework}.
The regression-based framework needs to consider the global shape of the text, so the robustness is higher than the other two frameworks. The segmentation-based framework focuses on local features to distinguish the foreground background, making it more sensitive to image corruptions. The region proposal-based framework is a two-stage algorithm, and each stage is affected by corruption, so the robustness is slightly worse than the segmentation-based framework.

\myParabullet{Performance $w.r.t$ Pre-training Data.}
As shown in Table \ref{table_pretrain_simple}, pre-training data improves both the performance and robustness. On the IC15 dataset, the pre-training data improves the performance significantly, but on the CTW dataset, the improvement is smaller. The reason is that the pre-training data and the IC15 dataset are mostly multi-directional texts, while the CTW dataset has many curved texts. In addition, MLT17 has a more remarkable improvement in model performance than SynthText, and SynthText has a more significant improvement in model robustness. This shows that using real data (MLT17) instead of synthetic data (SynthText) to pre-train the network can improve the model performance because synthetic data cannot fully simulate real data, and there is a gap with real data. Synthetic data is generally obtained by embedding text in a smooth area on the background image. The text can appear in a more diverse range of locations. The texture of the text area is more complex, and the network also has some robustness to image corruptions by learning this kind of data.

\begin{table}[t]
\centering
\small
\renewcommand{\arraystretch}{1.}
\renewcommand{\tabcolsep}{1.9pt}
\begin{tabular}{c|c|c||c|ccccc|c|c} 
\hline
$\mathcal{M}$                    & $\mathcal{D}$                    & $\mathcal{P}$  & \textbf{Clean} & \textbf{N.} & \textbf{B.}  & \textbf{W.} & \textbf{D.} & \textbf{G.} & \textbf{mPC} & \textbf{rPC} \\ 
\hline
\multirow{6}{*}{\rotatebox{90}{MSRCNN}} & \multirow{3}{*}{\rotatebox{90}{CTW}}   & None      & 73.2 & 56.4 & 31.9 & 69.5   & 56.3   & 50.6   & 52.5   & 71.7    \\
                          &                            & SynthText & 75.5 & \textbf{60.3} & 35.0 & \textbf{71.9}   & \textbf{58.6}   & \textbf{58.4}     & \textbf{55.8}   & \textbf{73.9}   \\
                          &                            & MLT17     & \textbf{75.6} & 59.2 & \textbf{37.7} & 71.7   & 57.8   & 56.6     & \textbf{55.8}   & 73.8   \\ 
\cline{2-11}
                          & \multirow{3}{*}{\rotatebox{90}{IC15}} & None      & 82.5 & 16.7 & 26.2 & 61.6   & 62.4   & 68.0     & 47.2   & 57.3   \\
                          &                            & SynthText & 84.5 & \textbf{20.6} & \textbf{27.3} & \textbf{63.6}   & 64.1   & \textbf{71.5}    & \textbf{68.6}   & \textbf{81.3}    \\
                          &                            & MLT17     & \textbf{85.4} & 16.8 & 27.2 & 62.7   & \textbf{64.8}   & \textbf{71.5}     & 68.1   & 80.5   \\ 
\hline
\multirow{6}{*}{\rotatebox{90}{FCENet}}   & \multirow{3}{*}{\rotatebox{90}{CTW}}   & None      & 84.3 & 72.5 & 40.9 & 82.1   & 68.8   & 74.5    & 66.1   & 78.4    \\
                          &                            & SynthText & 84.4 & \textbf{76.0} & 44.8 & \textbf{83.1}   & \textbf{71.1}   & \textbf{76.1}    & \textbf{68.6}   & \textbf{81.3}    \\
                          &                            & MLT17     & \textbf{84.6} & 74.7 & \textbf{45.8} & 82.9   & 69.8   & 75.5     & 68.1   & 80.5   \\ 
\cline{2-11}
                          & \multirow{3}{*}{\rotatebox{90}{IC15}} & None      & 84.9 & 19.6 & 30.1 & 65.9   & 67.3   & 76.8    & 51.9   & 61.1    \\
                          &                            & SynthText & 85.5 & \textbf{34.2} & \textbf{31.7} & \textbf{69.1}   & \textbf{69.0}   & \textbf{78.4}    & \textbf{56.0}   & \textbf{65.5}    \\
                          &                            & MLT17     & \textbf{85.6} & 22.5 & 31.3 & 66.8   & 68.2   & 77.1     & 53.1   & 62.0   \\ 
\hline
\multirow{6}{*}{\rotatebox{90}{PSENet}}   & \multirow{3}{*}{\rotatebox{90}{CTW}}   & None      & 78.8 & 57.0 & 29.7 & 76.1   & 59.5   & 56.0     & 54.8   & 69.6   \\
                          &                            & SynthText & \textbf{81.4} & \textbf{67.6} & 27.5 & \textbf{78.5}   & \textbf{62.3}   & \textbf{62.6}    & \textbf{58.2}   & 71.5    \\
                          &                            & MLT17     & 80.6 & 64.1 & \textbf{31.8} & 77.9   & 61.6   & 61.7    & 58.1   & \textbf{72.1}    \\ 
\cline{2-11}
                          & \multirow{3}{*}{\rotatebox{90}{IC15}} & None      & 80.7 & 21.6 & 26.9 & 59.5   & 63.2   & 67.8     & 48.1   & 59.6   \\
                          &                            & SynthText & 82.2 & \textbf{37.7} & 27.9 & \textbf{64.7}   & 65.9   & 69.9     & \textbf{53.0}   & \textbf{64.5}   \\
                          &                            & MLT17     & \textbf{83.6} & 33.6 & \textbf{30.0} & 63.2   & \textbf{66.4}   & \textbf{72.2}     & \textbf{53.0}   & 63.4   \\
\hline
\end{tabular}
\caption{Results of different pre-training data on CTW and IC15. ``None'' means no pre-training data is used.}
\label{table_pretrain_simple}
\end{table}

\myParabullet{Performance $w.r.t$ Backbone.}
\begin{table}[t]
\centering
\small
\renewcommand{\arraystretch}{1}
\renewcommand{\tabcolsep}{1.1pt}
\begin{tabular}{c|c|c||c|ccccc|c|c}
\hline
$\mathcal{M}$                    & $\mathcal{D}$                    & $\mathcal{B}$ & \textbf{Clean}  & \textbf{N.}  & \textbf{B.}   & \textbf{W.} & \textbf{D.} & \textbf{G.} & \textbf{mPC} & \textbf{rPC}  \\ 
\hline
\multirow{8}{*}{\rotatebox{90}{MSRCNN}} & \multirow{4}{*}{\rotatebox{90}{CTW}}   & VGG16    & 61.2~ & 43.4~ & 22.8~ & 54.9~  & 46.0~  & 37.8~  & 41.0  & 67.0   \\
                          &                            & ResNet50 & 73.2~ & 56.4~ & 31.9~ & 69.5~  & 56.3~  & 50.6~  & 52.5  & 71.7  \\
                          &                            & ResNeXt  & 74.3~ & 56.7~ & 33.1~ & 70.7~  & 56.7~  & 51.4~  & 53.2  & 71.6  \\
                          &                            & RegNet   & \textbf{76.1~} & \textbf{61.2} & \textbf{35.8} & \textbf{73.1}  & \textbf{60.3}  & \textbf{53.1}  & \textbf{56.3}  & \textbf{74.0}  \\ 
\cline{2-11}
                          & \multirow{4}{*}{\rotatebox{90}{IC15}} & VGG16    & 72.6~ & 8.9~ & 19.7~ & 41.7~  & 55.6~  & 50.5~  & 37.0  & 50.9  \\
                          &                            & ResNet50 & 82.5~ & 16.7~ & 26.2~ & 61.6~  & 62.4~  & 68.0~  & 47.2  & 57.3  \\
                          &                            & ResNeXt  & 83.9~ & 14.2~ & 27.3~ & 62.0~  & 64.7~  & 67.8~  & 47.9  & 57.1  \\
                          &                            & RegNet   & \textbf{84.1} & \textbf{20.9} & \textbf{27.9} & \textbf{63.8}  & \textbf{66.4}  & \textbf{69.2}  & \textbf{50.1}  & \textbf{59.6}  \\ 
\hline
\multirow{8}{*}{\rotatebox{90}{FCENet}}   & \multirow{4}{*}{\rotatebox{90}{CTW}}   & VGG16    & 65.6~ & 49.7~ & 23.4~ & 61.2~  & 50.6~  & 28.0~  & 43.6  & 66.5  \\
                          &                            & ResNet50 & 84.3~ & 72.5~ & 40.9~ & 82.1~  & 68.8~  & 74.5~  & 66.1  & 78.4  \\
                          &                            & ResNeXt  & 84.3~ & \textbf{75.6} & \textbf{43.6} & 82.4~  & \textbf{69.4}  & 73.8~  & \textbf{67.3}  & \textbf{79.9}  \\
                          &                            & RegNet   & \textbf{85.3} & 73.4~ & 41.2~ & \textbf{82.9}  & 68.8~  & \textbf{75.2}  & 66.5  & 77.9  \\ 
\cline{2-11}
                          & \multirow{4}{*}{\rotatebox{90}{IC15}} & VGG16    & 78.9~ & 20.0~ & 27.5~ & 62.6~  & 62.2~  & 68.5~  & 48.3  & 61.2  \\
                          &                            & ResNet50 & 84.9~ & 19.6~ & 30.1~ & 65.9~  & 67.3~  & 76.8~  & 51.9  & 61.1  \\
                          &                            & ResNeXt  & 84.5~ & \textbf{30.5} & 30.4~ & 66.1~  & 68.4~  & 76.4~  & 54.1  & 64.0  \\
                          &                            & RegNet   & \textbf{86.1} & 29.9~ & \textbf{31.0} & \textbf{68.2}  & \textbf{70.0}  & \textbf{78.4}  & \textbf{55.3}  & \textbf{64.2}  \\ 
\hline
\multirow{8}{*}{\rotatebox{90}{PSENet}}   & \multirow{4}{*}{\rotatebox{90}{CTW}}   & VGG16    & 75.4~ & 55.2~ & 20.8~ & 71.7~  & 53.1~  & 49.7~  & 49.0  & 65.0  \\
                          &                            & ResNet50 & 78.8~ & 57.0~ & 29.7~ & 76.1~  & 59.5~  & 56.0~  & 54.8  & 69.6  \\
                          &                            & ResNeXt  & 79.0~ & 63.6~ & 28.8~ & 77.2~  & 61.0~  & 58.4~  & 56.7  & 71.7  \\
                          &                            & RegNet   & \textbf{79.5} & \textbf{65.8} & \textbf{31.0} & \textbf{77.7}  & \textbf{61.1}  & \textbf{59.9}  & \textbf{57.8}  & \textbf{72.7}  \\ 
\cline{2-11}
                          & \multirow{4}{*}{\rotatebox{90}{IC15}} & VGG16    & 80.4~ & 26.2~ & 24.0~ & 60.6~  & 61.1~  & 67.0~  & 47.6  & 59.3  \\
                          &                            & ResNet50 & 80.7~ & 21.6~ & 26.9~ & 59.5~  & 63.2~  & 67.8     & 48.1   & 59.6  \\
                          &                            & ResNeXt  & 82.1~ & 26.4~ & 28.4~ & 64.0~  & 65.5~  & 70.1~  & 51.0  & 62.1  \\
                          &                            & RegNet   & \textbf{82.4} & \textbf{34.9} & \textbf{29.0} & \textbf{65.4}  & \textbf{66.0}  & \textbf{70.9~}  & \textbf{53.0}  & \textbf{64.3}  \\
\hline
\end{tabular}
\caption{ Results of different backbones on CTW and IC15.}
\label{table:backbone_simple}
\end{table}
As shown in Table \ref{table:backbone_simple}, RegNet achieves higher performance and robustness compared to other networks. We find that different backbone networks have less effect on segmentation-based PSENet, probably because such methods only need local texture features to distinguish text and non-text regions, so a general backbone network can achieve good results.
In addition, it is observed that a better backbone improves the robustness of the model on the noise the most, which indicates that a better feature extractor can better resist the effect of noise.

\myParabullet{Performance $w.r.t$ Feature Fusion.}
The results of feature fusion modules are shown in Table \ref{table:neck}. PAFPN incorporates shallow features into deep features, while CARAFE proposes a new feature upsampling operator with the advantage of a greater receptive field and the ability to generate the upsampling kernel based on the input features dynamically. PAFPN and CARAFE can improve the performance on clean images. PAFPN reduces the robustness of the model to noise, probably because of interference from shallow features. But CARAFE can slightly improve the robustness because it can focus on contextual information and is less susceptible to corruption. For PSENet, different feature fusion modules all have different effects on robustness, with FPNF being the most robust to geometric perturbations and FPN\_UNET being the most robust to noise.

\begin{table}[t]
\centering
\small
\renewcommand{\arraystretch}{1.}
\renewcommand{\tabcolsep}{1.4pt}
\begin{tabular}{c|c||c|ccccc|c|c} 
\hline
$\mathcal{M}$                                         & \textbf{Neck}      & \textbf{Clean} & \textbf{N.} & \textbf{B.} & \textbf{W.} & \textbf{D.} & \textbf{G.} & \textbf{mPC} & \textbf{rPC}  \\ 
\hline
\multicolumn{1}{l|}{\multirow{3}{*}{\rotatebox{45}{MSRCNN}}} & FPN       & 82.5 & 16.7 & 26.2                    & 61.6                       & 62.4   & 68.0    & 47.2   & 57.3                      \\
\multicolumn{1}{l|}{}                          & PAFPN     & \textbf{83.8} & 12.8 & \textbf{27.4}                    & 59.5                       & \textbf{63.3}   & \textbf{69.0}     & 46.9   & 56.0                     \\
\multicolumn{1}{l|}{}                          & CARAFE    & 83.0 & \textbf{22.4} & 26.8                   & \textbf{63.3}                       & 62.6   & 68.9            & \textbf{48.8}   & \textbf{58.7}                \\ 
\hline
\multicolumn{1}{l|}{\multirow{3}{*}{\rotatebox{45}{FCENet}}} & FPN       & 84.9 & 19.6 & 30.1                    & 65.9                       & 67.3   & 76.8                    & 51.9   & 61.1        \\
\multicolumn{1}{l|}{}                          & PAFPN     & \textbf{85.7} & 15.5 & 28.6                    & 66.1                       & 66.0   & \textbf{77.1}               & 50.5   & 58.9             \\
\multicolumn{1}{l|}{}                          & CARAFE    & 85.2 & \textbf{24.1} & \textbf{30.3}                    & \textbf{66.7}                       & \textbf{68.0}   & 76.4               & \textbf{53.0}   & \textbf{62.2}             \\ 
\hline
\multirow{4}{*}{\rotatebox{45}{PSENet}}                        & FPNF      & \textbf{80.7} & 21.6 & \textbf{26.9}                    & \textbf{59.5}                       & 63.2   & 67.8     & 48.1   & 59.6         \\
                                               & FFM       & 79.6 & 24.4 & 26.4                    & 58.9                       & \textbf{63.7}   & \textbf{68.2}                & \textbf{48.6}   & 61.0            \\
                                               & FPNC      & 78.5 & 24.1 & 25.2                    & 55.9                       & 61.7   & 63.9             & 47.9   & \textbf{61.1}               \\
                                               & FPN\_U & 78.5 & \textbf{26.2} & 24.3                    & 56.6                       & 62.4   & 63.3                & 47.0   & 60.0            \\
\hline
\end{tabular}
\caption{Results of different feature fusion modules on IC15, ``FPN\_U'' means ``FPN\_UNET''.}
\label{table:neck}
\end{table}
\begin{table}[t]
\centering
\small
\renewcommand{\arraystretch}{1.}
\renewcommand{\tabcolsep}{3.pt}
\begin{tabular}{ccc||c|ccccc|c|c} 
\hline
\textbf{P3} & \textbf{P4} & \textbf{P5} & \textbf{Clean} & \textbf{N.} & \textbf{B.}  & \textbf{W.} & \textbf{D.} & \textbf{G.}  & \textbf{mPC} & \textbf{rPC}   \\ 
\hline
\checkmark  & \checkmark  & \checkmark  & 84.9 & 19.6 & \textbf{30.1} & 65.9   & \textbf{67.3}   & 76.8    & \textbf{51.9}   & 61.1   \\
\checkmark  &    &    & \textbf{85.0} & 18.8 & \textbf{30.1} & \textbf{66.0}   & 66.6   & \textbf{77.0}     & 51.6   & 60.7    \\
   & \checkmark &    & 83.7 & 19.8 & 29.7 & 65.8   & 65.1   & 73.8     & 50.8   & 60.7    \\
   &    & \checkmark  & 72.0     & \textbf{23.9}     & 24.9     & 57.4       & 56.1       & 63.5      & 44.9   & \textbf{62.3}       \\
\hline
\end{tabular}
\caption{Results of FCENet with different scale features on IC15. ``P3'', ``P4'' and ``P5'' represent the features after $8 \times$, $16 \times$, and $32 \times$ downsampling, respectively.}
\label{table:multiscale}
\end{table}

\myParabullet{Performance $w.r.t$ Multi Scale Predictions.}
The results of multi-scale predictions are shown in Table \ref{table:multiscale}. This shows that using only P3 or P4 to predict is similar to using all scale features, while only P5 are much worse. We believe that in the IC15 dataset, the scale of text lines does not vary much, so it is sufficient to use only P3 or P4 features. Interestingly, the results for P5 correspond to better robustness in the case of noise than the others, probably because the features with a large receptive field are more robust to noise. Observe the results of different scale features for geometric 
corruptions: the F-scores of P3, P4, and P5 are 77.0, 73.8, and 63.5, respectively, and it can be found that the shallow features are more robust to geometric perturbations.

\myParabullet{Performance $w.r.t$ Representations of Text Instances.}
The results of different representations on CTW are shown in Table \ref{table:reprenstation}. For segmentation-based framework, PSENet outperforms other methods in the case of blur and noise; DRRG performs better in weather, digital and geometry situations; the robustness of PAN is the worst. PSENet considers the contextual information to predict text kernels at different scales. DRRG predicts a series of small text components, focusing more on local features, and thus is robust to geometric perturbations. But the noise and blur cause significant perturbation to the prediction of local components, so DRRG is sensitive to them. PAN predicts text regions to get the whole text shape and distinguishes different text instances by similarity vectors. When the image is corrupted, the prediction of both similarity vectors and text regions will be disturbed. Combining the wrong text regions with the wrong similarity vectors will get worse results.
Fourier-based and polar coordinates-based robustness for the regression-based framework is about the same. The former performs better on clean images because the Fourier signal can better represent curved text. They both predict parametric representations of text contours by neural networks. Specifically, both directly regress a series of values to represent a closed curve and have similar robustness.

\begin{table}[t]
\centering
\small
\renewcommand{\arraystretch}{1.}
\renewcommand{\tabcolsep}{2.4pt}
\begin{tabular}{c|c||c|ccccc|c|c} 
\hline
$\mathcal{M}$                                         & \textbf{Head}      & \textbf{Clean} & \textbf{N.} & \textbf{B.} & \textbf{W.} & \textbf{D.} & \textbf{G.} & \textbf{mPC} & \textbf{rPC}  \\ 
\hline
\multicolumn{1}{l|}{\multirow{3}{*}{Seg.}} & TK       & 77.2 & \textbf{61.0} & \textbf{29.6}                    & 75.6                       & 60.0   & 57.5    & 55.7   & \textbf{72.2}                      \\
\multicolumn{1}{l|}{}                          & CC     & \textbf{83.2} & 58.4 & 28.4                    & \textbf{78.6}                       & \textbf{61.1}   & \textbf{73.4}     & \textbf{57.7}   & 69.3                     \\
\multicolumn{1}{l|}{}                          & SV    & 77.3 & 54.7 & 20.6                    & 71.6                       & 56.2   & 61.2            & 51.1   & 66.2                \\ 
\hline
\multicolumn{1}{l|}{\multirow{2}{*}{Regr.}} & Fourier       & \textbf{84.3} & \textbf{72.5} & 40.9                   & \textbf{82.1}                       & \textbf{68.8}   & \textbf{74.5}                    & \textbf{66.1}   & 78.4        \\
\multicolumn{1}{l|}{}                          & Polar     & 82.7 & 72.4 & \textbf{41.1}                    & 80.7                       & 67.1   & 71.4               & 65.0   & \textbf{78.5}             \\
\hline
\end{tabular}
\caption{Representations of text instances. ``Seg.'' indicates the segmentation-based framework. ``Regr.'' indicates the regression-based framework. ``TK'', ``CC'' and ``SV'' denotes text kernel (PSENet), local connected components (DRRG) and similar vectors (PAN).}
\label{table:reprenstation}
\end{table}

\begin{table}[t]
\centering
\small
\renewcommand{\arraystretch}{1.}
\renewcommand{\tabcolsep}{3.pt}
\begin{tabular}{c||c|ccccc|c|c} 
\hline
\textbf{Loss}       & \textbf{Clean} & \textbf{N.} & \textbf{B.} & \textbf{W.}    & \textbf{D.} & \textbf{G.} & \textbf{mPC} & \textbf{rPC} \\ 
\hline
\textbf{L1}         & 82.5 & 16.7 & 26.2                    & 61.6                            & 62.4   & 68.0      & 47.2   & 57.3                     \\
\textbf{SmoothL1}   & \textbf{83.5} & 11.5 & \textbf{27.4}                    & 59.7                       & \textbf{63.4}   & 67.5           & 46.6   & 55.8                 \\
\textbf{IoU}        & 83.3 & \textbf{22.6} & 25.8                    & 62.1                       & 62.2   & 68.3            & \textbf{48.2}   & 57.8                \\
\textbf{GIoU}       & 82.8 & 21.5 & 26.1                    & 62.2                       & 62.4   & \textbf{68.4}            & \textbf{48.2}   & \textbf{58.2}                \\
\textbf{DIoU}    & 83.1 & 20.8 & 25.4                    & \textbf{62.5}                       & 62.1   & 68.0               & 47.8   & 57.5             \\
\hline
\end{tabular}
\caption{Results of different loss functions on IC15.}
\label{tabel:loss}
\end{table}

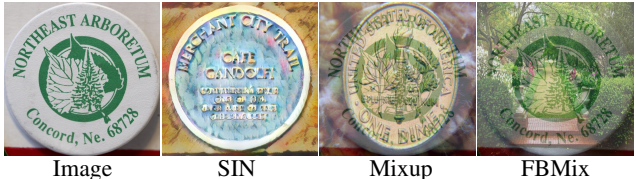
\begin{figure}[tb]
\centering
\begin{overpic}[width=1.0\linewidth]{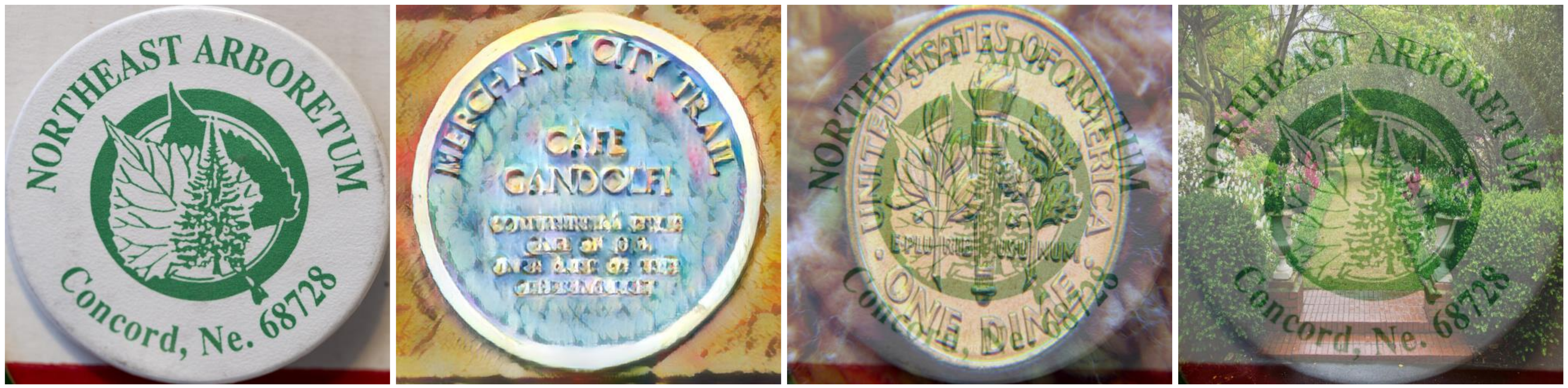}
            \put(8 , -3.0){\small{Image}}
            \put(34, -3.0){\small{SIN}}
            \put(58, -3.0){\small{Mixup}}
            \put(82, -3.0){\small{FBMix}}
\end{overpic}
\caption{Comparison of augmentation methods.}
\label{fig:aug}
\end{figure}

\begin{table}[h]
\centering
\small
\renewcommand{\arraystretch}{1.0}
\renewcommand{\tabcolsep}{2.pt}
\begin{tabular}{c|c|c||c|ccccc|c|c}
\hline
$\mathcal{M}$                    & $\mathcal{D}$                    & $\mathcal{A}$      & \textbf{Clean} & \textbf{N.} & \textbf{B.}  & \textbf{W.} & \textbf{D.} & \textbf{G.}  & \textbf{mPC} & \textbf{rPC} \\ 
\hline
\multirow{8}{*}{\rotatebox{90}{MSRCNN}} & \multirow{4}{*}{\rotatebox{90}{CTW}}   & Baseline & 73.2 & 56.4 & 26.2 & 69.5   & 56.3   & \textbf{50.6}    & 52.5   & 71.7   \\
                          &                            & SIN      & 71.5 & 61.3 & 29.3 & 69.7   & 58.5   & 49.0     & 55.8   & \textbf{78.0}   \\
                          &                            & MixUp    & 72.1 & 57.9 & 33.9 & 69.6   & \textbf{59.5}   & 47.1     & 55.3   & 76.7    \\ 
                          &                            & FBMix  & \textbf{74.0} & \textbf{62.7} & \textbf{35.1} & \textbf{71.2}   & 59.4   & 50.0    & \textbf{56.2}   & 76.0     \\
\cline{2-11}
                          & \multirow{4}{*}{\rotatebox{90}{IC15}} & Baseline & 82.5 & 16.7 & 26.2 & 61.6   & 62.4   & \textbf{68.0}     & 47.2   & 57.3    \\
                          &                            & SIN      & 80.3 & 42.6 & 29.3 & 61.2   & 65.1   & 60.8     & 52.3   & 65.1    \\
                          &                            & MixUp    & \textbf{83.5} & 22.1 & 33.9 & 67.9   & 67.3   & 64.5     & 52.1   & 62.4    \\ 
                          &                            & FBMix  & 83.0 & \textbf{55.8} & \textbf{35.1} & \textbf{68.2}   & \textbf{68.2}   & 63.4     & \textbf{58.2}   & \textbf{70.1}    \\
\hline
\multirow{8}{*}{\rotatebox{90}{FCENet}}   & \multirow{4}{*}{\rotatebox{90}{CTW}}   & Baseline & 84.3 & 72.5 & 40.9 & 82.1   & 68.8   & 74.5    & 66.1   & 78.4     \\
                          &                            & SIN      & 84.4 & 76.2 & 51.8 & 82.8   & 71.1   & \textbf{76.2}    & 70.2   & 83.2     \\
                          &                            & MixUp    & 85.0 & 79.2 & 52.8 & 83.3   & 74.3   & 72.6     & 71.6   & 84.3    \\ 
                          &                            & FBMix  & \textbf{85.9} & \textbf{81.4} & \textbf{60.3} & \textbf{85.0}   & \textbf{75.3}   & 75.9     & \textbf{74.7}   & \textbf{87.0}    \\
\cline{2-11}
                          & \multirow{4}{*}{\rotatebox{90}{IC15}} & Baseline & 84.9 & 19.6 & 30.1 & 65.9   & 67.3   & \textbf{76.8}    & 51.9   & 61.1     \\
                          &                            & SIN      & 82.5 & 35.7 & 38.8 & 67.1   & 67.7   & 73.2     & 56.4   & 68.4    \\
                          &                            & MixUp    & \textbf{85.0} & 43.5 & 41.6 & 72.8   & 69.7   & 75.4    & 60.2   & 70.9     \\ 
                          &                            & FBMix  & 84.7 & \textbf{55.1} & \textbf{44.2} & \textbf{73.8}   & \textbf{70.3}   & 73.9    & \textbf{62.9}   & \textbf{74.3}     \\
\hline
\multirow{8}{*}{\rotatebox{90}{PSENet}}   & \multirow{4}{*}{\rotatebox{90}{CTW}}   & Baseline & 78.8 & 57.0 & 29.7 & 76.1   & 59.5   & 56.0     & 54.8   & 69.6    \\
                          &                            & SIN      & 79.3 & 68.8 & 33.1 & 77.7   & 62.4   & \textbf{60.7}    & 59.3   & 74.8     \\
                          &                            & MixUp    & 79.7 & \textbf{69.6} & \textbf{37.5} & 78.5   & \textbf{64.7}   & 60.2     & \textbf{61.2}   & \textbf{76.9}    \\ 
                          &                            & FBMix  & \textbf{80.3} & 68.7 & 36.1 & \textbf{79.6}   & \textbf{64.7}   & 59.3     & 60.9   & 75.9    \\
\cline{2-11}
                          & \multirow{4}{*}{\rotatebox{90}{IC15}} & Baseline & \textbf{80.7} & 21.6 & 26.9 & 59.5   & 63.2   & \textbf{67.8}     & 48.1   & 59.6    \\
                          &                            & SIN      & 75.2 & 33.3 & 24.4 & 53.9   & 59.4   & 59.2    & 46.3   & 61.6     \\
                          &                            & MixUp    & 78.1 & 35.4 & \textbf{32.0} & \textbf{63.4}   & \textbf{64.8}   & 63.5     & 52.2   & 66.9    \\
                          &                            & FBMix  & 78.0 & \textbf{45.6} & 30.5 & 62.8   & 63.5   & 61.4     & \textbf{52.8}   & \textbf{67.7}    \\
\hline
\end{tabular}
\caption{Comparisons of different data augmentation methods on CTW and IC15.}
\label{table:aug}
\end{table}

\myParabullet{Performance $w.r.t$ Loss Function.}
The results of MSRCNN trained on the IC15 dataset using different loss functions are shown in Table \ref{tabel:loss}, which shows that the IOU-based loss function is more robust to noise and can improve performance. For other types of image corruptions, the L1-based and IOU-based loss functions perform similarly.

\subsection{Improving Robustness by FBMix}

Comparisons of FBMix with other data augmentations are shown in Fig. \ref{fig:aug}. SIN stylizes the input image to destroy the texture information but potentially compromises the structural information of the text. MixUp blends the two foreground images without introducing additional texture information and with possible overlap between the text. FBMix blends a foreground image with a natural background image without text, introducing more background texture information in the text region.

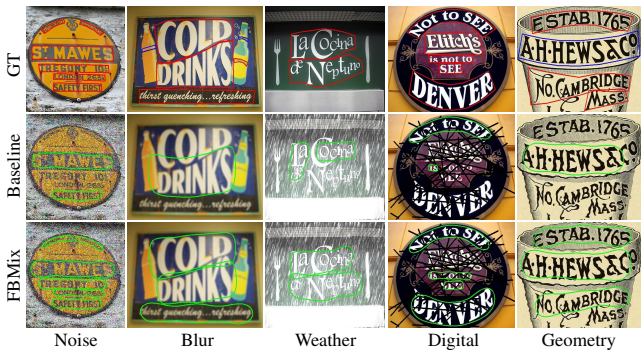
\begin{figure}[h!]
\centering
\scriptsize
\begin{overpic}[width=1.0\linewidth]{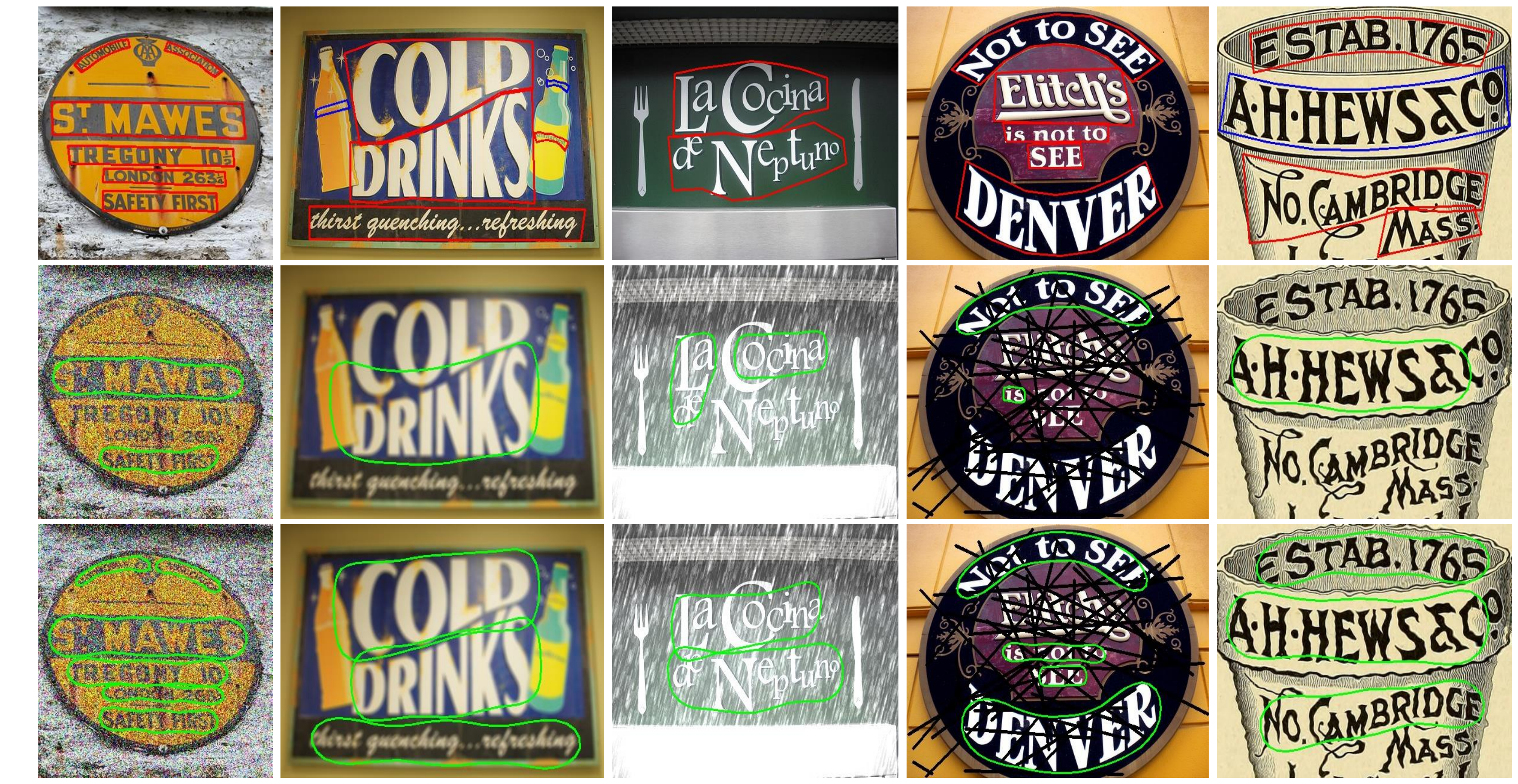}
    \put(-0.5, 40){\rotatebox{90}{GT}}
    \put(-0.5, 21.){\rotatebox{90}{Baseline}}
    \put(-0.5, 4.5){\rotatebox{90}{FBMix}}
    
    \put(7, -2.5){Noise}
    \put(27, -2.5){Blur}
    \put(45, -2.5){Weather}
    \put(66, -2.5){Digital}
    \put(84, -2.5){Geometry}
\end{overpic}
\caption{Qualitative comparisons between FCENet without and
with FBMix.}
\label{fig:result}
\end{figure}

\myPara{Qualitative Results.} We illustrate several challenging results and make some visual comparisons in Fig. \ref{fig:result}. The results in the first row indicate the original image and its corresponding annotation, and the blue box means the ignorable text. The results in the second row show that even if the text on the image is visible after the corruption, the model may still predict incorrectly. The comparisons demonstrate that FBMix mitigates the interference caused by image corruptions, significantly improves the robustness.

\myPara{Quantitative Results.} The results of data augmentations are shown in Table \ref{table:aug}. Although the SIN can increase the robustness, it has a likelihood to turn the text areas into background texture, which will introduce label noise during the training process. Causally, the performance of the model on clean images will degrade significantly. FBMix outperforms SIN and MixUp in most cases and has the largest improvement in robustness to noisy and blurred data. MixUp fuses two images containing text in the training dataset without introducing richer background textures, so the performance is worse than FBMix in most cases.

\section{Conclusion}
We conduct a detailed comparison of scene text detection robustness against eighteen image corruptions. We systematically study the various influencing factors that affect the robustness of text detection, including text detection framework, pre-training data, backbone, feature fusion module, multi-scale predictions, representation of text instances, and loss function. Based on the experimental results, we obtained some important conclusions: better backbone network can improve the performance and robustness of the model; using global information to model text contours helps to improve the overall robustness of the model; shallow features are sensitive to noise; IoU-based loss is more effective than L1-based loss. 
We find that synthetic data improves the robustness of the model more than real data because the background texture in the text area of synthetic data is more diverse. We accordingly propose a simple but effective data augmentation method FBMix, which significantly improves the robustness of three existing text detection frameworks and maintains the performance. These findings are crucial for researchers when designing models for text detection tasks. And we hope that this work will draw the community's attention to the robustness of text detection.

\bibliography{aaai22}

\clearpage
\section{Supplemental Material}
In this supplementary material, we provide additional results to complement the paper. First, we provide more detailed experimental results to complement the article to further illustrate the aspects that affect the robustness of scene text detection. Second, we show more qualitative results.

\begin{table*}
\centering
\scriptsize
\renewcommand{\arraystretch}{1.}
\renewcommand{\tabcolsep}{2.5pt}
\begin{tabular}{c|c|c|c|ccc|cccc|ccc|cccccc|cc} 
\hline
                        &                       &           &                & \multicolumn{3}{c|}{Noise}                       & \multicolumn{4}{c|}{Blur}                                         & \multicolumn{3}{c|}{Weather}                     & \multicolumn{6}{c|}{Digital}                                                                        & \multicolumn{2}{c}{Geometry}     \\ 
\hline
$\mathcal{M}$                  & $\mathcal{D}$               & Pretrain  & Clean          & Gauss       & Shot           & Impulse        & Defocus        & Glass          & Motion         & Zoom           & Snow           & Frost          & Fog            & Bright     & Contrast       & Pixel       & Jpeg           & Dirty          & Lines          & Rotation       & Elastic         \\ 
\hline
\multirow{6}{*}{\rotatebox{90}{MSRCNN}} & \multirow{3}{*}{\rotatebox{90}{CTW}}  & None      & 73.2~          & 56.9~          & 56.7~          & 55.7~          & 30.6~          & 32.3~          & 33.3~          & 31.5~          & 67.1~          & 69.9~          & 71.6~          & 72.6~          & 61.8~          & 53.8~          & 61.2~          & 57.9~          & 30.7~          & 41.1~          & 60.1~           \\
                        &                       & SynthText & 75.5~          & {60.6~} & {60.6~} & {59.8~} & 32.5~          & 36.6~          & 35.5~          & 35.3~          & {69.2~} & {72.5~} & {74.0~} & 74.0~          & {65.7~} & {56.8~} & 63.5~          & {61.4~} & 29.8~          & {52.5~} & {64.3~}  \\
                        &                       & MLT17     & {75.6~} & 59.7~          & 59.5~          & 58.6~          & {34.2~} & {41.1~} & {39.4~} & {36.2~} & {69.2~} & 72.1~          & 73.7~          & {74.2~} & 63.2~          & 53.7~          & {63.9~} & 59.3~          & {32.7~} & 48.8~          & {64.3~}  \\ 
\cline{2-22}
                        & \multirow{3}{*}{\rotatebox{90}{IC15}} & None      & 82.5~          & 18.5~          & 16.7~          & 14.8~          & 28.7~          & 29.5~          & 38.2~          & {8.5~}  & 46.5~          & 62.6~          & 75.8~          & 76.6~          & 64.7~          & 48.5~          & 66.5~          & 54.9~          & 63.6~          & 55.5~          & 80.5~           \\
                        &                       & SynthText & 84.5~          & {22.1~} & {20.4~} & {19.2~} & 28.6~          & {32.2~} & {41.1~} & 7.5~           & {48.2~} & {65.1~} & 77.5~          & {78.6~} & {67.5~} & 46.3~          & 70.7~          & 54.0~          & 67.6~          & {61.3~} & 81.7~           \\
                        &                       & MLT17     & {85.4~} & 18.8~          & 16.7~          & 14.9~          & {29.2~} & 30.9~          & 40.3~          & {8.4~}  & 46.5~          & 64.1~          & {77.6~} & 78.1~          & 67.4~          & {48.8~} & {70.8~} & {55.5~} & {68.2~} & 60.9~          & {82.1~}  \\ 
\hline
\multirow{6}{*}{\rotatebox{90}{FCENet}} & \multirow{3}{*}{\rotatebox{90}{CTW}}  & None      & 84.3~          & 73.2~          & 72.7~          & 71.5~          & 36.8~          & 43.9~          & 45.5~          & 37.4~          & 80.3~          & 81.9~          & 84.1~          & 83.2~          & 80.7~          & 66.0~          & 73.5~          & 68.9~          & 40.4~          & 76.2~          & 72.9~           \\
                        &                       & SynthText & 84.4~          & {75.8~} & {76.4~} & {75.8~} & 39.7~          & 47.8~          & 49.5~          & 42.1~          & {81.6~} & {83.3~} & 84.4~          & 84.0~          & {82.4~} & {69.1~} & {77.1~} & 71.1~          & {42.8~} & {76.8~} & {75.3~}  \\
                        &                       & MLT17     & {84.6~} & 75.1~          & 75.1~          & 73.9~          & {41.0~} & {48.3~} & {51.7~} & {42.4~} & 81.1~          & 82.9~          & {84.8~} & {84.4~} & 81.7~          & 67.4~          & 75.5~          & {71.9~} & 37.9~          & 75.9~          & 75.0~           \\ 
\cline{2-22}
                        & \multirow{3}{*}{\rotatebox{90}{IC15}} & None      & 84.9~          & 22.7~          & 21.9~          & 14.1~          & 29.0~          & 34.2~          & 43.4~          & 13.6~          & 47.1~          & 66.6~          & 84.0~          & 79.8~          & 79.1~          & 56.9~          & 65.5~          & 57.1~          & 65.7~          & 70.6~          & 83.1~           \\
                        &                       & SynthText & 85.5~          & {35.6~} & {36.5~} & {30.4~} & 30.3~          & {36.7~} & {45.5~} & {14.4~} & {52.9~} & {69.7~} & {84.6~} & {82.1~} & {80.9~} & 57.0~          & {67.5~} & 58.6~          & {67.8~} & {72.4~} & {84.4~}  \\
                        &                       & MLT17     & {85.6~} & 25.4~          & 24.8~          & 17.2~          & {30.5~} & 36.4~          & 45.1~          & 12.9~          & 48.7~          & 67.6~          & 84.2~          & 81.5~          & 79.6~          & {57.7~} & 63.9~          & {60.3~} & 66.2~          & 70.4~          & 83.7~           \\ 
\hline
\multirow{6}{*}{\rotatebox{90}{PSENet}} & \multirow{3}{*}{\rotatebox{90}{CTW}}  & None      & 78.8~          & 57.9~          & 57.7~          & 55.5~          & 30.8~          & 33.1~          & 30.4~          & 24.5~          & 74.3~          & 76.1~          & 77.8~          & 77.5~          & 67.1~          & 57.2~          & 65.7~          & 61.7~          & 28.0~          & 56.9~          & 55.1~           \\
                        &                       & SynthText & {81.4~} & {68.2~} & {67.8~} & {66.7~} & 27.8~          & 33.2~          & 25.5~          & 23.5~          & 75.7~          & {78.9~} & {81.1~} & {80.0~} & {72.5~} & {58.5~} & {69.6~} & {65.9~} & 27.1~          & 60.1~          & {65.2~}  \\
                        &                       & MLT17     & 80.6~          & 65.2~          & 65.0~          & 62.0~          & {31.5~} & {36.2~} & {32.0~} & {27.7~} & {76.5~} & 78.0~          & 79.2~          & 79.4~          & 71.1~          & 58.3~          & 68.8~          & 63.0~          & {29.4~} & {62.9~} & 60.6~           \\ 
\cline{2-22}
                        & \multirow{3}{*}{\rotatebox{90}{IC15}} & None      & 80.7~          & 24.6~          & 22.8~          & 17.5~          & 27.5~          & 33.9~          & 38.1~          & 8.2~           & 45.7~          & 59.3~          & 73.5~          & 75.4~          & 61.5~          & 60.0~          & 68.9~          & 53.4~          & 60.1~          & 58.2~          & 77.4~           \\
                        &                       & SynthText & 82.2~          & {39.8~} & {36.7~} & {36.4~} & 28.1~          & 34.3~          & 36.5~          & {13.0~} & {51.2~} & {65.1~} & {77.9~} & 78.5~          & {66.8~} & 60.1~          & {70.5~} & 55.1~          & {64.6~} & 60.4~          & 79.4~           \\
                        &                       & MLT17     & {83.6~} & 37.4~          & 34.7~          & 28.9~          & {31.4~} & {37.2~} & {40.3~} & 11.3~          & 47.6~          & 64.6~          & 77.5~          & {78.8~} & 65.8~          & {62.3~} & {70.5~} & {55.6~} & 65.3~          & {62.9~} & {81.5~}  \\
\hline
\end{tabular}
\caption{Results of pre-training data on CTW and IC15.}
\label{table:pretrain_d}
\end{table*}

\subsection{1.More detailed results}
In this section, we provide more detailed results.

\myParabullet{Performance $w.r.t$ Pre-training Data.}
The results of different pre-training data are shown in Table \ref{table:pretrain_d}. The experimental results show that SynText mainly improves the robustness of the model to various types of noise. This suggests that improving the diversity of background textures in the text area helps to combat noise. We observe that the improvement of SynthText's robustness against noise is much more significant than MLT17 with the IC15 dataset. This is because IC15 contains a lot of text at more minor scales, which is challenging to be detected correctly under various kinds of noise. In contrast, the text in CTW is at a larger scale, and the noise does not affect it too much. That is, the greater the perturbation caused by the noise to the text in the image, the more significant the improvement of SynthText.

\myParabullet{Performance $w.r.t$ Backbone.}
The results of different backbones are shown in Table \ref{table:backbone_d}. The effect of the backbone network on model robustness is all-encompassing, i.e., it affects the performance of the model to various types of corruptions simultaneously. This suggests that a better feature extractor provides more resistant features to perturbations and, therefore, can improve the robustness to different types of perturbations.

\myParabullet{Performance $w.r.t$ Feature Fusion.}
The results of feature fusion modules are shown in Table \ref{table:neck_d}. For region proposal-based and regression-based methods, the different feature fusion modules mainly affect the robustness of the model to noise and blur. For segmentation-based methods, there is no apparent advantage or disadvantage of different feature fusion modules.

\myParabullet{Performance $w.r.t$ Multi Scale Predictions.}
The results of multi-scale predictions are shown in Table \ref{table:ms_d}. This table contains the results of each type of corruption compared to the article. 

\myParabullet{Performance $w.r.t$ Representations of Text Instances.}
The results of different representations on CTW are shown in Table \ref{table:rep_d}. This table contains the results of each type of corruption compared to the article.

\myParabullet{Performance $w.r.t$ Loss Function.}
The results of MSRCNN trained on the IC15 dataset using different loss functions are shown in Table \ref{table:loss_d}. This table contains the results of each type of corruption compared to the article.

\myParabullet{Performance $w.r.t$ Data Augmentations.}
The results of data augmentations are shown in Table \ref{table:aug_d}. SIN outperforms other methods in improving the robustness of the model to blur under the CTW dataset. Because the process of stylizing the image by SIN changes the structural information of the text, and the scale of the text in CTW is large, the text is not completely transformed into the background texture, and part of the structure is distorted but still visible as text, similar to the blurring effect. In addition, for the most part, several of the data augmentation methods reduce the robustness of the model to rotation, probably because they all tend to make the model ignore texture information. Hence, the model focuses on local text structure feature learning and ignores global information. And when the image is rotated, the local structural features of the text change a lot, so the performance is poor.

\subsection{2.More qualitative results}
In this section, we show more qualitative results.

\begin{figure*}[ht]
\centering
\begin{overpic}[width=1.0\linewidth]{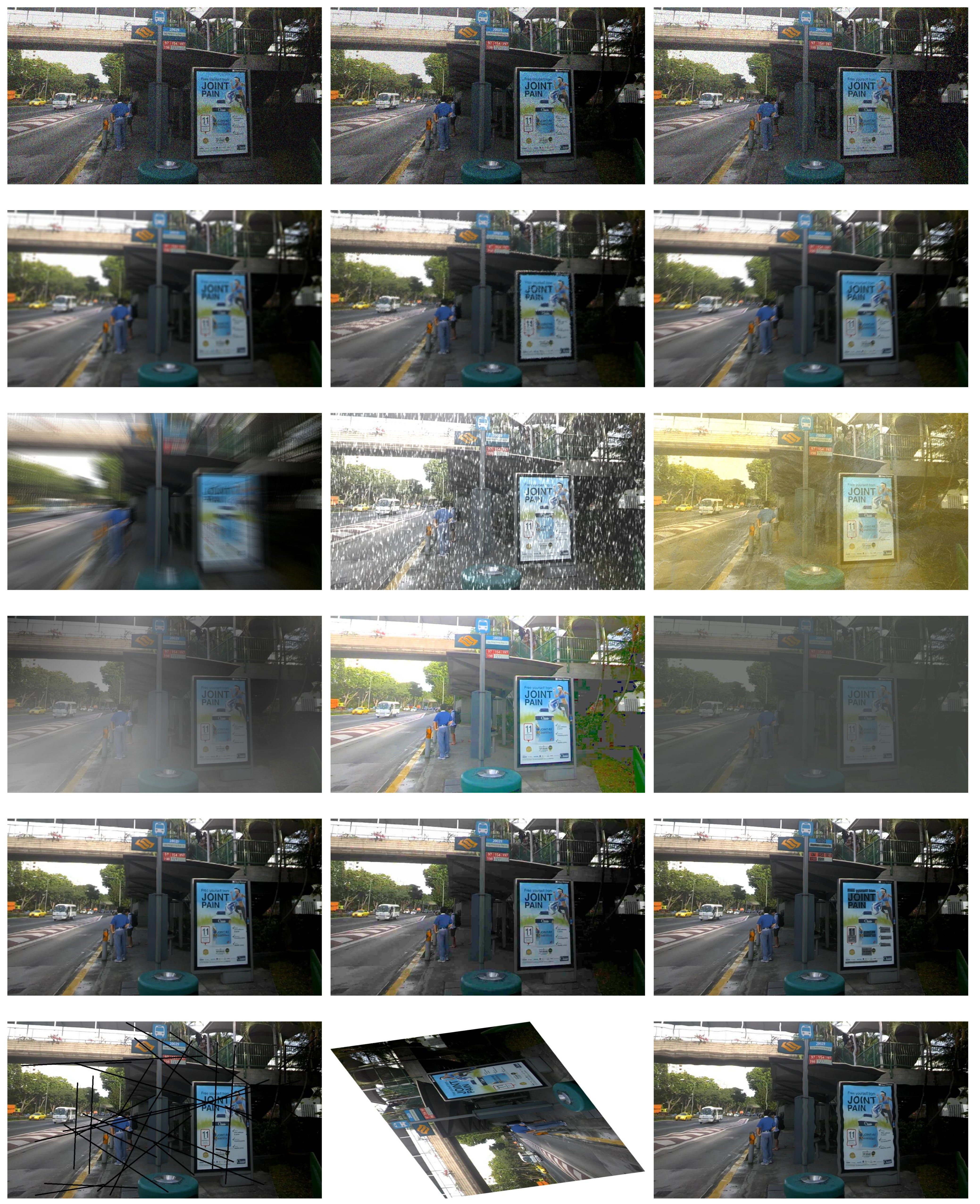}
            \put(8, 83){{Gaussian noise}}
            \put(37, 83){{Shot noise}}
            \put(63, 83){{Impulse noise}}
            \put(8, 66.5){{Defocus blur}}
            \put(36, 66.5){{Glass blur}}
            \put(63, 66.5){{Motion blur}}
            \put(10, 49.5){{Zoom blur}}
            \put(38, 49.5){{Snow}}
            \put(65, 49.5){{ Frost}}
            \put(12, 32.5){{Fog}}
            \put(37, 32.5){{Brightness}}
            \put(65, 32.5){{Contrast}}
            \put(11, 15.7){{Pixelate}}
            \put(39, 15.7){{Jpeg}}
            \put(66, 15.7){{Dirty}}
            \put(11, -1){{Lines}}
            \put(38, -1){{Rotation}}
            \put(65, -1){{Elastic}}
		    
\end{overpic}
\caption{Visualization of images from our benchmark datasets, which  consists of 18 types of corruptions from noise, blur, weather, digital and geometry categories. There are five levels of severity for each type of corruption, leading to 90 distinct corruptions.}
\label{fig:ic15}
\end{figure*}

\myParabullet{Benchmark Datasets}
The example of CTW-C is shown in the article, here we add the example of IC15-C, as shown in Fig. \ref{fig:ic15}.

\begin{figure*}[ht]
\centering
\begin{overpic}[width=1.0\linewidth]{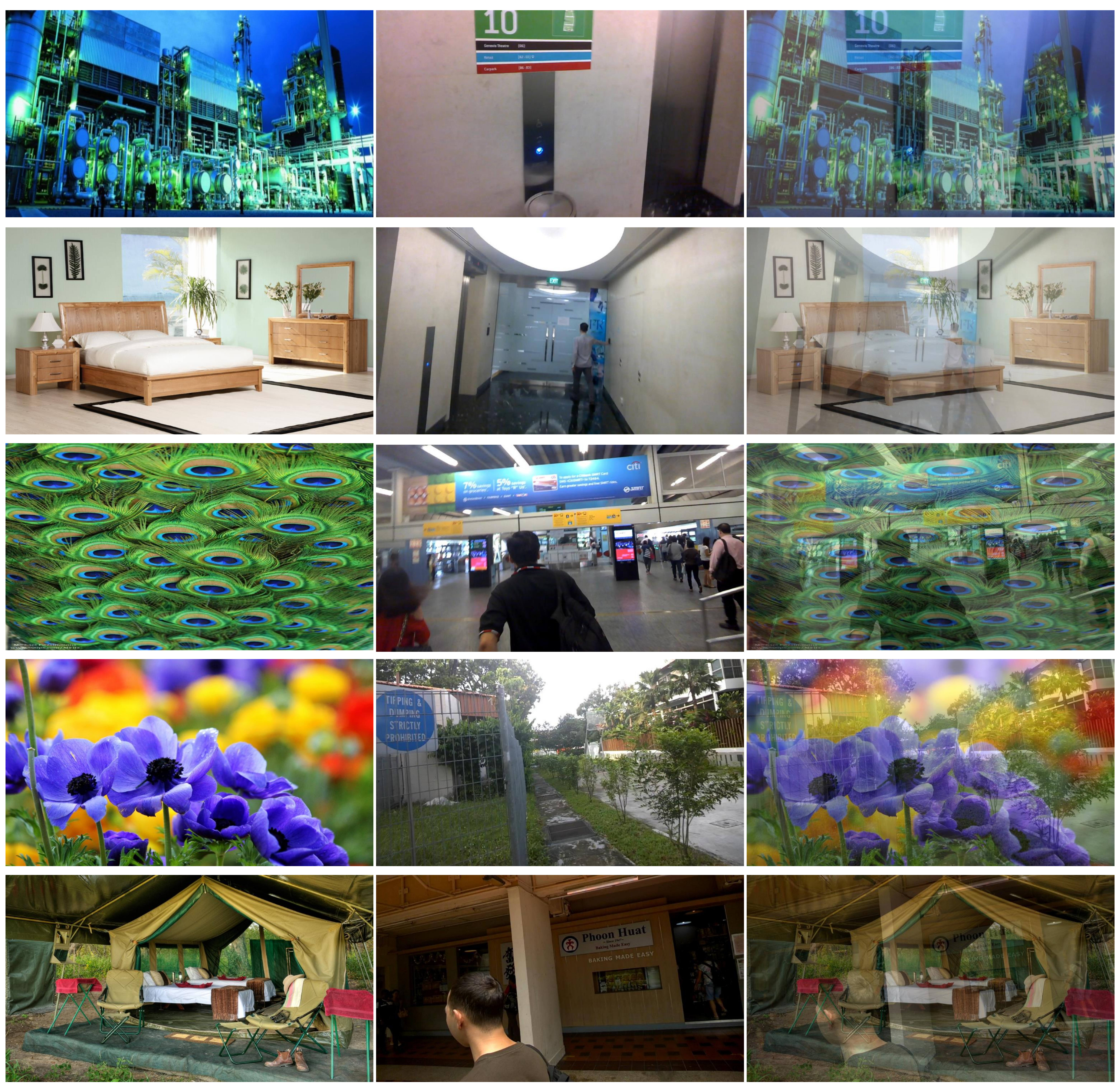}
            \put(13, -1.5){{Background}}
            \put(48, -1.5){{Foreground}}
            \put(80, -1.5){{FBMix}}
		    
\end{overpic}
\caption{Realizations of FBMix, combine foreground image and background image together to get a new image.}
\label{fig:fbmix}
\end{figure*}

\myParabullet{Realizations of FBMix}
Example images of FBMix are shown in Fig. \ref{fig:fbmix}. FBMix greatly improves the variety of background textures in the text area by introducing images from natural scenes.

\begin{figure*}[ht]
\centering
\begin{overpic}[width=0.8\linewidth]{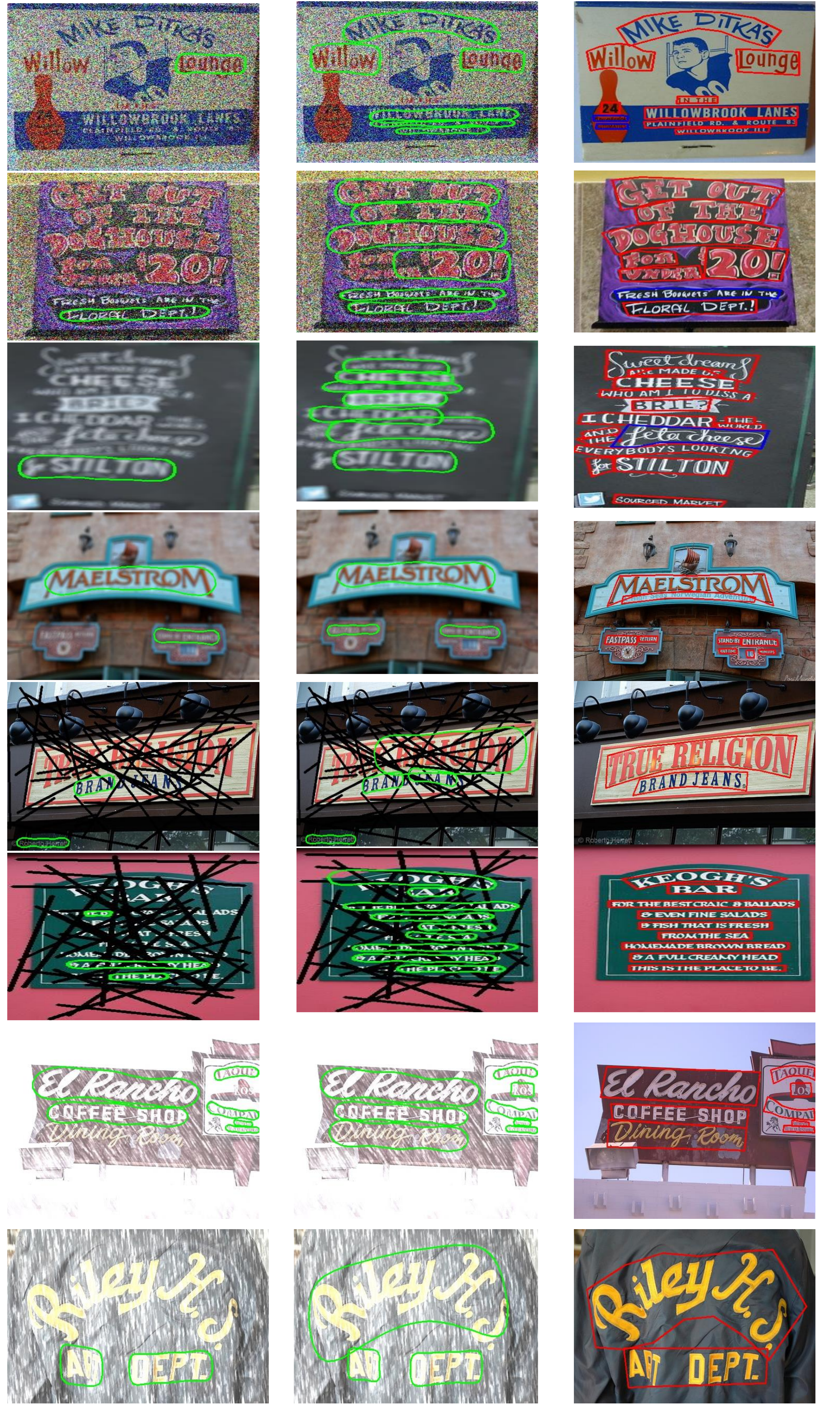}
            \put(7, 0){{w/o FBMix}}
            \put(26,0){{w/ FBMix}}
            \put(47,0){{GT}}
		    
\end{overpic}
\caption{Qualitative comparisons between without and with FBMix.}
\label{fig:augres}
\end{figure*}

\myParabullet{Effect of FBMix}
The results of using FBMix to train the model are shown in Fig. \ref{fig:augres}.  The results demonstrate that FBMix mitigates the interference caused by image corruptions, significantly improves the robustness.

\begin{table*}
\centering
\scriptsize
\renewcommand{\arraystretch}{1.}
\renewcommand{\tabcolsep}{2.5pt}
\begin{tabular}{c|c|c|c|ccc|cccc|ccc|cccccc|cc} 
\hline
                          &                            &          &       & \multicolumn{3}{c|}{Noise} & \multicolumn{4}{c|}{Blur}        & \multicolumn{3}{c|}{Weather} & \multicolumn{6}{c|}{Digital}                      & \multicolumn{2}{c}{Geometry}  \\ 
\hline
$\mathcal{M}$                    & $\mathcal{D}$                     & $\mathcal{B}$  & Clean & Gauss & Shot  & Impulse    & Defocus & Glass & Motion & Zoom  & Snow  & Frost & Fog          & Bright & Contrast & Pixel & Jpeg  & Dirty & Lines & Rotation   & Elastic               \\ 
\hline
\multirow{8}{*}{\rotatebox{90}{MSRCNN}} & \multirow{4}{*}{\rotatebox{90}{CTW}}   & VGG16    & 61.2~ & 44.0~ & 44.3~ & 41.8~      & 20.3~   & 22.8~ & 23.0~  & 25.1~ & 51.2~ & 55.6~ & 58.0~        & 60.1~  & 43.0~    & 49.0~ & 54.2~ & 44.6~ & 25.3~ & 24.9~ & 50.7~                 \\
                          &                            & ResNet50 & 73.2~ & 56.9~ & 56.7~ & 55.7~      & 30.6~   & 32.3~ & 33.3~  & 31.5~ & 67.1~ & 69.9~ & 71.6~        & 72.6~  & 61.8~    & 53.8~ & 61.2~ & 57.9~ & 30.7~ & 41.1~ & 60.1~                 \\
                          &                            & ResNeXt  & 74.3~ & 56.6~ & 56.4~ & 57.3~      & 31.4~   & 36.3~ & 33.7~  & 30.7~ & 67.7~ & 71.2~ & 73.0~        & 73.0~  & 63.4~    & 50.8~ & 61.2~ & 58.7~ & 33.0~ & 41.3~ & 61.4~                 \\
                          &                            & RegNet   & 76.1~ & 61.3~ & 61.4~ & 60.9~      & 32.8~   & 36.9~ & 38.5~  & 34.9~ & 71.1~ & 73.5~ & 74.8~        & 74.7~  & 65.8~    & 58.7~ & 65.2~ & 63.5~ & 33.6~ & 42.6~ & 63.5~                 \\
            \cline{2-22}
                          & \multirow{4}{*}{\rotatebox{90}{IC15}} & VGG16    & 72.6~ & 7.9~  & 9.6~  & 9.1~       & 20.3~   & 25.3~ & 28.1~  & 5.1~  & 22.3~ & 42.1~ & 60.7~        & 67.0~  & 46.7~    & 56.0~ & 64.8~ & 41.6~ & 57.6~ & 31.0~ & 70.1~                 \\
                          &                            & ResNet50 & 82.5~ & 18.5~ & 16.7~ & 14.8~      & 28.7~   & 29.5~ & 38.2~  & 8.5~  & 46.5~ & 62.6~ & 75.8~        & 76.6~  & 64.7~    & 48.5~ & 66.5~ & 54.9~ & 63.6~ & 55.5~ & 80.5~                 \\
                          &                            & ResNeXt  & 83.9~ & 16.1~ & 14.5~ & 12.0~      & 28.5~   & 32.9~ & 39.8~  & 8.0~  & 44.2~ & 63.2~ & 78.5~        & 77.9~  & 67.2~    & 51.2~ & 69.1~ & 55.9~ & 67.1~ & 54.6~ & 80.9~                 \\
                          &                            & RegNet   & 84.1~ & 20.5~ & 19.3~ & 22.9~      & 29.4~   & 32.5~ & 40.8~  & 8.9~  & 49.9~ & 64.4~ & 77.0~        & 78.0~  & 64.7~    & 58.1~ & 71.3~ & 57.6~ & 68.6~ & 57.0~ & 81.3~                 \\ 
\hline
\multirow{8}{*}{\rotatebox{90}{FCENet}}   & \multirow{4}{*}{\rotatebox{90}{CTW}}   & VGG16    & 65.6~ & 50.3~ & 50.6~ & 48.1~      & 18.2~   & 25.2~ & 25.0~  & 25.1~ & 58.0~ & 61.0~ & 64.5~        & 63.8~  & 56.9~    & 48.5~ & 57.6~ & 49.7~ & 27.0~ & 0.9~  & 55.2~                 \\
                          &                            & ResNet50 & 84.3~ & 73.2~ & 72.7~ & 71.5~      & 36.8~   & 43.9~ & 45.5~  & 37.4~ & 80.3~ & 81.9~ & 84.1~        & 83.2~  & 80.7~    & 66.0~ & 73.5~ & 68.9~ & 40.4~ & 76.2~ & 72.9~                 \\
                          &                            & ResNeXt  & 84.3~ & 76.0~ & 76.1~ & 74.6~      & 41.2~   & 45.5~ & 49.6~  & 38.3~ & 80.9~ & 82.4~ & 83.8~        & 83.3~  & 81.2~    & 62.8~ & 74.5~ & 72.7~ & 41.9~ & 76.2~ & 71.4~                 \\
                          &                            & RegNet   & 85.3~ & 73.5~ & 73.5~ & 73.3~      & 38.8~   & 46.4~ & 47.1~  & 32.5~ & 81.5~ & 82.6~ & 84.4~        & 84.2~  & 78.6~    & 67.2~ & 75.6~ & 70.1~ & 37.2~ & 76.8~ & 73.5~                 \\
                    \cline{2-22}
                          & \multirow{4}{*}{\rotatebox{90}{IC15}} & VGG16    & 78.9~ & 21.0~ & 19.9~ & 19.1~      & 24.9~   & 29.2~ & 39.3~  & 16.8~ & 47.6~ & 63.0~ & 77.4~        & 74.5~  & 71.4~    & 48.0~ & 62.7~ & 55.2~ & 61.7~ & 59.7~ & 77.4~                 \\
                          &                            & ResNet50 & 84.9~ & 22.7~ & 21.9~ & 14.1~      & 29.0~   & 34.2~ & 43.4~  & 13.6~ & 47.1~ & 66.6~ & 84.0~        & 79.8~  & 79.1~    & 56.9~ & 65.5~ & 57.1~ & 65.7~ & 70.6~ & 83.1~                 \\
                          &                            & ResNeXt  & 84.5~ & 34.5~ & 33.6~ & 23.3~      & 30.7~   & 35.9~ & 42.9~  & 12.1~ & 48.4~ & 66.4~ & 83.3~        & 81.3~  & 79.0~    & 58.5~ & 66.1~ & 59.4~ & 66.2~ & 69.7~ & 83.1~                 \\
                          &                            & RegNet   & 86.1~ & 31.8~ & 31.9~ & 26.0~      & 31.0~   & 37.0~ & 44.0~  & 12.0~ & 51.7~ & 69.0~ & 83.9~        & 81.0~  & 79.8~    & 59.6~ & 68.4~ & 63.6~ & 67.6~ & 72.5~ & 84.2~                 \\ 
\hline
\multirow{8}{*}{\rotatebox{90}{PSENet}}   & \multirow{4}{*}{\rotatebox{90}{CTW}}   & VGG16    & 75.4~ & 56.5~ & 56.1~ & 53.1~      & 22.6~   & 25.9~ & 19.8~  & 14.9~ & 69.5~ & 71.7~ & 74.0~        & 74.7~  & 63.6~    & 37.8~ & 57.3~ & 57.3~ & 27.8~ & 53.0~ & 46.4~                 \\
                          &                            & ResNet50 & 78.8~ & 57.9~ & 57.7~ & 55.5~      & 30.8~   & 33.1~ & 30.4~  & 24.5~ & 74.3~ & 76.1~ & 77.8~        & 77.5~  & 67.1~    & 57.2~ & 65.7~ & 61.7~ & 28.0~ & 56.9~ & 55.1~                 \\
                          &                            & ResNeXt  & 79.0~ & 64.3~ & 64.2~ & 62.2~      & 31.1~   & 33.3~ & 27.5~  & 23.3~ & 75.1~ & 77.3~ & 79.3~        & 78.2~  & 70.8~    & 56.7~ & 65.2~ & 65.3~ & 29.8~ & 57.4~ & 59.4~                 \\
                          &                            & RegNet   & 79.5~ & 66.0~ & 66.1~ & 65.4~      & 33.9~   & 34.5~ & 29.6~  & 25.8~ & 76.4~ & 77.8~ & 78.8~        & 78.9~  & 71.9~    & 51.2~ & 68.1~ & 64.8~ & 32.0~ & 60.3~ & 59.4~                 \\
                 \cline{2-22}
                          & \multirow{4}{*}{\rotatebox{90}{IC15}} & VGG16    & 80.4~ & 28.2~ & 24.3~ & 26.3~      & 26.0~   & 26.9~ & 36.2~  & 7.1~  & 47.5~ & 59.5~ & 74.9~        & 74.6~  & 62.6~    & 48.6~ & 66.3~ & 52.6~ & 62.2~ & 56.4~ & 77.5~                 \\
                          &                            & ResNet50 & 80.7~ & 24.6~ & 22.8~ & 17.5~      & 27.5~   & 33.9~ & 38.1~  & 8.2~  & 45.7~ & 59.3~ & 73.5~        & 75.4~  & 61.5~    & 60.0~ & 68.9~ & 53.4~ & 60.1~ & 58.2~ & 77.4~                 \\
                          &                            & ResNeXt  & 82.1~ & 30.5~ & 27.3~ & 21.3~      & 28.9~   & 34.9~ & 38.7~  & 10.9~ & 51.0~ & 63.1~ & 78.1~        & 77.8~  & 66.9~    & 60.7~ & 70.0~ & 55.2~ & 62.5~ & 60.9~ & 79.4~                 \\
                          &                            & RegNet   & 82.4~ & 36.2~ & 34.2~ & 34.2~      & 30.4~   & 34.1~ & 39.6~  & 11.8~ & 54.0~ & 64.0~ & 78.1~        & 78.9~  & 66.9~    & 59.0~ & 70.8~ & 56.1~ & 64.1~ & 61.4~ & 80.4~                 \\
\hline
\end{tabular}
\caption{Results of backbones on CTW and IC15.}
\label{table:backbone_d}
\end{table*}

\begin{table*}
\centering
\scriptsize
\renewcommand{\arraystretch}{1.}
\renewcommand{\tabcolsep}{2.5pt}
\begin{tabular}{c|c|c|ccc|cccc|ccc|cccccc|cc} 
\hline
                          &        & \multicolumn{1}{l|}{}      &  \multicolumn{3}{c|}{Noise} & \multicolumn{4}{c|}{Blur}        & \multicolumn{3}{c|}{Weather} & \multicolumn{6}{c|}{Digital}                      & \multicolumn{2}{c}{Geometry}  \\ 
\hline
Method                   & Neck   & \multicolumn{1}{l|}{Clean} & \multicolumn{1}{l}{Gauss} & \multicolumn{1}{l}{Shot} & \multicolumn{1}{l|}{Impulse} & \multicolumn{1}{l}{Defocus} & \multicolumn{1}{l}{Glass} & \multicolumn{1}{l}{Motion} & \multicolumn{1}{l|}{Zoom} & \multicolumn{1}{l}{Snow} & \multicolumn{1}{l}{Frost} & \multicolumn{1}{l|}{Fog} & \multicolumn{1}{l}{Bright} & \multicolumn{1}{l}{Contrast} & \multicolumn{1}{l}{Pixel} & \multicolumn{1}{l}{Jpeg} & \multicolumn{1}{l}{Dirty} & \multicolumn{1}{l|}{Lines} & \multicolumn{1}{l}{Rotation} & \multicolumn{1}{l}{Elastic}  \\ 
\hline
\multirow{3}{*}{\rotatebox{45}{MSRCNN}} & FPN    & 82.5~ & 18.5~ & 16.7~ & 14.8~      & 28.7~   & 29.5~ & 38.2~  & 8.5~  & 46.5~ & 62.6~ & 75.8~        & 76.6~  & 64.7~    & 48.5~ & 66.5~ & 54.9~ & 63.6~ & 55.5~ & 80.5~                 \\
                          & PAFPN  & 83.8~ & 14.4~ & 12.4~ & 11.5~      & 29.7~   & 31.5~ & 40.5~  & 8.1~  & 42.9~ & 60.5~ & 75.0~        & 77.5~  & 63.5~    & 49.5~ & 69.0~ & 53.9~ & 66.4~ & 57.4~ & 80.6~                 \\
                          & CARAFE & 83.0~ & 24.2~ & 22.7~ & 20.4~      & 29.3~   & 30.6~ & 38.9~  & 8.4~  & 49.7~ & 63.8~ & 76.5~        & 76.6~  & 64.7~    & 47.3~ & 68.7~ & 55.3~ & 62.9~ & 57.4~ & 80.4~                 \\ 
\hline
\multirow{3}{*}{\rotatebox{45}{FCENet}}   & FPN    & 84.9~ & 22.7~ & 21.9~ & 14.1~      & 29.0~   & 34.2~ & 43.4~  & 13.6~ & 47.1~ & 66.6~ & 84.0~        & 79.8~  & 79.1~    & 56.9~ & 65.5~ & 57.1~ & 65.7~ & 70.6~ & 83.1~                 \\
                          & PAFPN  & 85.7~ & 18.3~ & 17.5~ & 10.8~      & 28.9~   & 32.9~ & 41.7~  & 11.0~ & 47.7~ & 66.9~ & 83.8~        & 80.3~  & 78.8~    & 50.5~ & 63.9~ & 58.1~ & 64.2~ & 70.3~ & 83.8~                 \\
                          & CARAFE & 85.2~ & 27.9~ & 26.7~ & 17.6~      & 30.8~   & 36.7~ & 42.5~  & 11.1~ & 48.7~ & 67.7~ & 83.7~        & 81.1~  & 80.0~    & 57.9~ & 65.1~ & 58.5~ & 65.3~ & 69.3~ & 83.5~                 \\ 
\hline
\multirow{4}{*}{\rotatebox{45}{PSENet}}   & FPNF   & 80.7~ & 24.6~ & 22.8~ & 17.5~      & 27.5~   & 33.9~ & 38.1~  & 8.2~  & 45.7~ & 59.3~ & 73.5~        & 75.4~  & 61.5~    & 60.0~ & 68.9~ & 53.4~ & 60.1~ & 58.2~ & 77.4~                 \\
                          & FFM    & 79.6~ & 26.4~ & 25.8~ & 20.9~      & 25.6~   & 33.1~ & 36.7~  & 10.3~ & 44.4~ & 58.3~ & 74.1~        & 74.4~  & 63.2~    & 62.8~ & 67.6~ & 51.6~ & 62.7~ & 59.6~ & 76.9~                 \\
                          & FPNC   & 78.5~ & 27.2~ & 25.3~ & 21.0~      & 25.6~   & 31.2~ & 34.9~  & 9.2~  & 41.7~ & 54.6~ & 71.3~        & 73.5~  & 58.0~    & 62.1~ & 66.5~ & 48.6~ & 61.6~ & 52.2~ & 75.7~                 \\
                          & FPNU   & 78.5~ & 28.0~ & 27.4~ & 23.2~      & 22.9~   & 31.8~ & 34.4~  & 8.0~  & 41.7~ & 55.5~ & 72.7~        & 73.6~  & 60.6~    & 62.3~ & 66.6~ & 49.2~ & 62.3~ & 50.3~ & 76.2~                 \\
\hline
\end{tabular}
\caption{Results of different feature fusion modules on IC15, ``FPN\_U'' means ``FPN\_UNET''.}
\label{table:neck_d}
\end{table*}

\begin{table*}
\centering
\scriptsize
\renewcommand{\arraystretch}{1.}
\renewcommand{\tabcolsep}{2.5pt}
\begin{tabular}{ccc|c|ccc|cccc|ccc|cccccc|cc} 
\hline
   &    &    &       & \multicolumn{3}{c|}{Noise} & \multicolumn{4}{c|}{Blur}        & \multicolumn{3}{c|}{Weather} & \multicolumn{6}{c|}{Digital}                      & \multicolumn{2}{c}{Geometry}  \\ 
\hline
P3 & P4 & P5 & Clean & Gauss & Shot  & Impulse    & Defocus & Glass & Motion & Zoom  & Snow  & Frost & Fog          & Bright & Contrast & Pixel & Jpeg  & Dirty & Lines & Rotation   & Elastic               \\
\hline
\checkmark  & \checkmark   & \checkmark   & 84.9~ & 22.7~ & 21.9~ & 14.1~      & 29.0~   & 34.2~ & 43.4~  & 13.6~ & 47.1~ & 66.6~ & 84.0~        & 79.8~  & 79.1~    & 56.9~ & 65.5~ & 57.1~ & 65.7~ & 70.6~ & 83.1~                 \\
\checkmark   &    &    & 85.0~ & 21.7~ & 21.2~ & 13.5~      & 29.9~   & 34.0~ & 44.2~  & 12.4~ & 46.5~ & 67.5~ & 83.9~        & 80.7~  & 77.9~    & 51.5~ & 65.2~ & 59.3~ & 64.9~ & 70.7~ & 83.3~                 \\
   & \checkmark   &    & 83.7~ & 23.2~ & 21.0~ & 15.3~      & 30.8~   & 34.2~ & 42.1~  & 11.8~ & 49.3~ & 66.7~ & 81.4~        & 78.4~  & 76.4~    & 49.8~ & 64.4~ & 58.9~ & 62.9~ & 65.8~ & 81.8~                 \\
   &    & \checkmark   & 72.0~ & 27.9~ & 27.3~ & 16.5~      & 27.1~   & 26.6~ & 36.0~  & 10.0~ & 43.7~ & 57.9~ & 70.7~        & 68.0~  & 66.0~    & 44.8~ & 53.9~ & 49.4~ & 54.5~ & 55.4~ & 71.6~                 \\
\hline
\end{tabular}
\caption{Results of FCENet with different scale features on IC15. ``P3'', ``P4'' and ``P5'' represent the features after $8 \times$, $16 \times$, and $32 \times$ downsampling, respectively.}
\label{table:ms_d}
\end{table*}

\begin{table*}
\centering
\scriptsize
\renewcommand{\arraystretch}{1.}
\renewcommand{\tabcolsep}{2.5pt}
\begin{tabular}{c|c|c|ccc|cccc|ccc|cccccc|cc} 
\hline
                       &         &       & \multicolumn{3}{c|}{Noise} & \multicolumn{4}{c|}{Blur}        & \multicolumn{3}{c|}{Weather} & \multicolumn{6}{c|}{Digital}                      & \multicolumn{2}{c}{Geometry}  \\ 
\hline
Method                      & Head    & Clean & Gauss & Shot  & Impulse    & Defocus & Glass & Motion & Zoom  & Snow  & Frost & Fog          & Bright & Contrast & Pixel & Jpeg  & Dirty & Lines & Rotation   & Elastic               \\ 
\hline
\multirow{3}{*}{Segmentation}  & TK      & 77.2~ & 62.1~ & 61.1~ & 59.8~      & 30.2~   & 32.3~ & 30.5~  & 25.4~ & 73.6~ & 75.7~ & 77.6~        & 76.9~  & 67.0~    & 58.8~ & 68.2~ & 60.8~ & 28.1~ & 57.9~ & 57.1~                 \\
                       & CC      & 83.2~ & 60.7~ & 58.4~ & 56.0~      & 24.3~   & 31.8~ & 32.4~  & 25.2~ & 76.1~ & 78.8~ & 80.9~        & 81.3~  & 64.0~    & 60.1~ & 68.8~ & 60.6~ & 31.7~ & 77.8~ & 68.9~                 \\
                       & SV      & 77.3~ & 57.5~ & 54.7~ & 51.8~      & 18.4~   & 26.0~ & 21.1~  & 16.7~ & 68.2~ & 71.2~ & 75.3~        & 75.9~  & 50.7~    & 60.3~ & 66.3~ & 55.2~ & 28.8~ & 58.2~ & 64.2~                 \\ 
\hline
\multirow{2}{*}{Regression} & Fourier & 84.3~ & 73.2~ & 72.7~ & 71.5~      & 36.8~   & 43.9~ & 45.5~  & 37.4~ & 80.3~ & 81.9~ & 84.1~        & 83.2~  & 80.7~    & 66.0~ & 73.5~ & 68.9~ & 40.4~ & 76.2~ & 72.9~                 \\
                       & Polar   & 82.7~ & 73.1~ & 73.1~ & 70.9~      & 36.9~   & 42.0~ & 46.3~  & 39.4~ & 79.1~ & 80.4~ & 82.6~        & 81.7~  & 79.5~    & 64.7~ & 72.8~ & 68.1~ & 36.0~ & 72.0~ & 70.8~                 \\
\hline
\end{tabular}
\caption{Representations of text instances. ``TK'', ``CC'' and ``SV'' denotes text kernel (PSENet), local connected components (DRRG) and similar vectors (PAN).}
\label{table:rep_d}
\end{table*}

\begin{table*}
\centering
\scriptsize
\renewcommand{\arraystretch}{1.}
\renewcommand{\tabcolsep}{2.5pt}
\begin{tabular}{c|c|ccc|cccc|ccc|cccccc|cc} 
\hline
         &       & \multicolumn{3}{c|}{Noise} & \multicolumn{4}{c|}{Blur}       & \multicolumn{3}{c|}{Weather} & \multicolumn{6}{c|}{Digital}                      & \multicolumn{2}{c}{Geometry}  \\ 
\hline
Loss     & Clean & Gauss & Shot  & Impulse    & Defocus & Glass & Motion & Zoom & Snow  & Frost & Fog          & Bright & Contrast & Pixel & Jpeg  & Dirty & Lines & Rotation   & Elastic               \\
\hline
L1       & 82.5~ & 18.5~ & 16.7~ & 14.8~      & 28.7~   & 29.5~ & 38.2~  & 8.5~ & 46.5~ & 62.6~ & 75.8~        & 76.6~  & 64.7~    & 48.5~ & 66.5~ & 54.9~ & 63.6~ & 55.5~ & 80.5~                 \\
SmoothL1 & 83.5~ & 13.1~ & 11.2~ & 10.2~      & 29.0~   & 30.5~ & 40.8~  & 9.1~ & 42.8~ & 61.4~ & 75.0~        & 76.9~  & 65.6~    & 49.6~ & 68.9~ & 53.0~ & 66.3~ & 54.4~ & 80.6~                 \\
IoU      & 83.3~ & 24.2~ & 22.5~ & 21.1~      & 29.0~   & 28.1~ & 38.1~  & 8.0~ & 48.0~ & 62.6~ & 75.9~        & 75.8~  & 64.6~    & 46.7~ & 67.4~ & 54.9~ & 63.9~ & 56.1~ & 80.5~                 \\
GIoU     & 82.8~ & 23.6~ & 21.3~ & 19.7~      & 29.7~   & 29.0~ & 38.4~  & 7.3~ & 48.6~ & 61.9~ & 75.9~        & 76.1~  & 65.2~    & 47.4~ & 67.5~ & 55.7~ & 62.6~ & 56.7~ & 80.1~                 \\
DIoU     & 83.1~ & 22.7~ & 21.0~ & 18.6~      & 28.0~   & 28.0~ & 37.7~  & 8.0~ & 48.4~ & 63.4~ & 75.6~        & 75.6~  & 64.7~    & 46.1~ & 67.6~ & 55.3~ & 63.5~ & 55.2~ & 80.8~                 \\
\hline
\end{tabular}
\caption{Results of different loss functions on IC15.}
\label{table:loss_d}
\end{table*}

\begin{table*}
\centering
\scriptsize
\renewcommand{\arraystretch}{1.}
\renewcommand{\tabcolsep}{2.5pt}
\begin{tabular}{c|c|c|c|ccc|cccc|ccc|cccccc|cc} 
\hline
                        &                            &          &       & \multicolumn{3}{c|}{Noise} & \multicolumn{4}{c|}{Blur}        & \multicolumn{3}{c|}{Weather} & \multicolumn{6}{c|}{Digital}                      & \multicolumn{2}{c}{Geometry}  \\ 
\hline
$\mathcal{M}$                 & $\mathcal{D}$                    & Aug      & Clean & Gauss & Shot  & Impulse    & Defocus & Glass & Motion & Zoom  & Snow  & Frost & Fog          & Bright & Contrast & Pixel & Jpeg  & Dirty & Lines & Rotation   & Elastic               \\ 
\hline
\multirow{8}{*}{\rotatebox{90}{MSRCNN}} & \multirow{4}{*}{\rotatebox{90}{CTW}}       & BaseLine & 73.2~ & 56.9~ & 56.7~ & 55.7~      & 30.6~   & 32.3~ & 33.3~  & 31.5~ & 67.1~ & 69.9~ & 71.6~        & 72.6~  & 61.8~    & 53.8~ & 61.2~ & 57.9~ & 30.7~ & 41.1~ & 60.1~                 \\
                        &                            & FBMix  & 74.0~ & 63.1~ & 62.0~ & 62.8~      & 33.6~   & 37.9~ & 40.5~  & 41.6~ & 69.4~ & 71.5~ & 72.8~        & 72.5~  & 68.0~    & 59.7~ & 63.4~ & 56.9~ & 35.6~ & 37.3~ & 62.8~                 \\
                        &                            & SIN      & 71.5~ & 61.4~ & 61.6~ & 61.0~      & 38.4~   & 39.1~ & 44.5~  & 40.0~ & 67.9~ & 69.8~ & 71.6~        & 70.9~  & 67.4~    & 55.3~ & 64.3~ & 62.6~ & 30.6~ & 35.3~ & 62.8~                 \\
                        &                            & MixUp    & 72.1~ & 58.3~ & 57.6~ & 57.8~      & 35.4~   & 37.6~ & 45.6~  & 43.1~ & 67.6~ & 69.8~ & 71.5~        & 70.9~  & 68.3~    & 54.4~ & 61.5~ & 60.9~ & 40.7~ & 35.2~ & 58.9~                 \\ 
\cline{2-22}
                        & \multirow{4}{*}{\rotatebox{90}{IC15}}      & BaseLine & 82.5~ & 18.5~ & 16.7~ & 14.8~      & 28.7~   & 29.5~ & 38.2~  & 8.5~  & 46.5~ & 62.6~ & 75.8~        & 76.6~  & 64.7~    & 48.5~ & 66.5~ & 54.9~ & 63.6~ & 55.5~ & 80.5~                 \\
                        &                            & FBMix  & 83.0~ & 55.6~ & 56.6~ & 55.1~      & 32.8~   & 37.4~ & 48.9~  & 21.3~ & 54.8~ & 69.6~ & 80.1~        & 79.2~  & 73.0~    & 57.6~ & 70.9~ & 58.4~ & 70.1~ & 45.9~ & 80.8~                 \\
                        &                            & SIN      & 80.3~ & 44.4~ & 44.8~ & 38.7~      & 27.0~   & 37.3~ & 40.2~  & 12.5~ & 46.8~ & 62.7~ & 74.2~        & 75.9~  & 66.1~    & 62.4~ & 70.6~ & 52.5~ & 63.3~ & 44.0~ & 77.5~                 \\
                        &                            & MixUp    & 83.5~ & 24.3~ & 22.2~ & 19.8~      & 29.9~   & 33.8~ & 49.4~  & 22.6~ & 53.5~ & 69.4~ & 80.9~        & 78.4~  & 75.1~    & 52.4~ & 69.2~ & 58.5~ & 70.0~ & 47.8~ & 81.1~                 \\ 
\hline
\multirow{8}{*}{\rotatebox{90}{FCENet}} & \multirow{4}{*}{\rotatebox{90}{CTW}}       & BaseLine & 84.3~ & 73.2~ & 72.7~ & 71.5~      & 36.8~   & 43.9~ & 45.5~  & 37.4~ & 80.3~ & 81.9~ & 84.1~        & 83.2~  & 80.7~    & 66.0~ & 73.5~ & 68.9~ & 40.4~ & 76.2~ & 72.9~                 \\
                        &                            & FBMix  & 85.9~ & 81.7~ & 81.4~ & 81.2~      & 52.0~   & 56.0~ & 68.7~  & 64.6~ & 84.0~ & 85.3~ & 85.7~        & 85.3~  & 83.9~    & 69.2~ & 77.6~ & 77.4~ & 58.6~ & 76.5~ & 75.3~                 \\
                        &                            & SIN      & 84.4~ & 76.5~ & 76.4~ & 75.7~      & 46.1~   & 52.6~ & 61.6~  & 47.1~ & 81.7~ & 82.7~ & 84.0~        & 83.7~  & 81.9~    & 69.6~ & 77.1~ & 71.7~ & 42.8~ & 76.3~ & 76.1~                 \\
                        &                            & MixUp    & 85.0~ & 79.2~ & 79.0~ & 79.3~      & 41.6~   & 48.2~ & 63.3~  & 58.2~ & 81.7~ & 83.6~ & 84.5~        & 83.9~  & 82.9~    & 69.9~ & 77.6~ & 77.6~ & 53.9~ & 69.8~ & 75.4~                 \\ 
\cline{2-22}
                        & \multirow{4}{*}{\rotatebox{90}{IC15}} & BaseLine & 84.9~ & 22.7~ & 21.9~ & 14.1~      & 29.0~   & 34.2~ & 43.4~  & 13.6~ & 47.1~ & 66.6~ & 84.0~        & 79.8~  & 79.1~    & 56.9~ & 65.5~ & 57.1~ & 65.7~ & 70.6~ & 83.1~                 \\
                        &                            & FBMix  & 84.7~ & 55.4~ & 54.6~ & 55.4~      & 38.7~   & 42.7~ & 58.8~  & 36.5~ & 61.3~ & 75.9~ & 84.0~        & 81.9~  & 81.0~    & 55.5~ & 66.4~ & 65.0~ & 71.8~ & 65.7~ & 82.1~                 \\
                        &                            & SIN      & 82.5~ & 38.6~ & 39.1~ & 29.4~      & 36.6~   & 47.9~ & 51.6~  & 18.9~ & 52.6~ & 67.4~ & 81.3~        & 79.0~  & 76.4~    & 60.4~ & 68.6~ & 57.1~ & 64.4~ & 65.6~ & 80.7~                 \\
                        &                            & MixUp    & 85.0~ & 44.4~ & 43.5~ & 42.7~      & 34.1~   & 41.2~ & 59.3~  & 31.9~ & 61.7~ & 72.4~ & 84.3~        & 80.9~  & 81.2~    & 57.1~ & 62.0~ & 65.6~ & 71.2~ & 67.9~ & 82.9~                 \\ 
\hline
\multirow{8}{*}{\rotatebox{90}{PSENet}} & \multirow{4}{*}{\rotatebox{90}{CTW}}   & BaseLine & 78.8~ & 57.9~ & 57.7~ & 55.5~      & 30.8~   & 33.1~ & 30.4~  & 24.5~ & 74.3~ & 76.1~ & 77.8~        & 77.5~  & 67.1~    & 57.2~ & 65.7~ & 61.7~ & 28.0~ & 56.9~ & 55.1~                 \\
                        &                            & FBMix  & 80.3~ & 69.9~ & 68.5~ & 67.7~      & 30.8~   & 33.9~ & 37.8~  & 41.8~ & 77.9~ & 79.6~ & 81.3~        & 79.3~  & 78.4~    & 58.7~ & 70.8~ & 65.4~ & 35.6~ & 58.6~ & 60.0~                 \\
                        &                            & SIN      & 79.3~ & 69.0~ & 68.8~ & 68.6~      & 31.6~   & 35.6~ & 34.5~  & 30.6~ & 76.5~ & 77.3~ & 79.3~        & 78.4~  & 71.8~    & 61.5~ & 71.3~ & 64.1~ & 27.4~ & 58.2~ & 63.2~                 \\
                        &                            & MixUp    & 79.7~ & 70.3~ & 69.0~ & 69.5~      & 30.8~   & 35.9~ & 43.4~  & 39.8~ & 76.1~ & 78.6~ & 80.7~        & 78.8~  & 77.7~    & 59.2~ & 71.5~ & 66.5~ & 34.2~ & 60.1~ & 60.2~                 \\ 
\cline{2-22}
                        & \multirow{4}{*}{\rotatebox{90}{IC15}} & BaseLine & 80.7~ & 24.6~ & 22.8~ & 17.5~      & 27.5~   & 33.9~ & 38.1~  & 8.2~  & 45.7~ & 59.3~ & 73.5~        & 75.4~  & 61.5~    & 60.0~ & 68.9~ & 53.4~ & 60.1~ & 58.2~ & 77.4~                 \\
                        &                            & FBMix  & 78.0~ & 46.2~ & 46.4~ & 44.2~      & 26.0~   & 35.2~ & 41.2~  & 19.7~ & 48.1~ & 64.5~ & 75.7~        & 74.7~  & 65.5~    & 58.6~ & 66.5~ & 52.4~ & 63.4~ & 46.9~ & 75.8~                 \\
                        &                            & SIN      & 75.2~ & 35.3~ & 36.2~ & 28.3~      & 23.7~   & 33.4~ & 31.2~  & 9.1~  & 39.6~ & 53.7~ & 68.4~        & 70.8~  & 57.4~    & 62.4~ & 63.2~ & 45.9~ & 57.0~ & 46.0~ & 72.3~                 \\
                        &                            & MixUp    & 78.1~ & 36.0~ & 35.2~ & 35.1~      & 27.6~   & 36.4~ & 43.7~  & 20.5~ & 50.4~ & 62.6~ & 77.3~        & 74.1~  & 70.4~    & 60.0~ & 66.2~ & 54.8~ & 63.1~ & 51.0~ & 76.1~                 \\
\hline
\end{tabular}
\caption{Results of different augmentations on CTW and IC15.}
\label{table:aug_d}
\end{table*}
\end{document}